\documentclass[conference]{IEEEtran}
\usepackage{times}
\usepackage{threeparttable}

\usepackage[numbers]{natbib}
\usepackage{multicol}
\newcommand{\tabincell}[2]{\begin{tabular}{@{}#1@{}}#2\end{tabular}}
\usepackage{color,xcolor}
\usepackage{epsfig}
\usepackage{graphicx}

\usepackage{adjustbox}
\usepackage{array}
\usepackage{booktabs}
\usepackage{colortbl}
\usepackage{float,wrapfig}
\usepackage{hhline}
\usepackage{multirow}
\let\labelindent\relax
\usepackage{enumitem}
\usepackage{amsmath,amsfonts,amsthm,amssymb}
\usepackage{bm}
\usepackage{nicefrac}
\usepackage{microtype}

\usepackage{changepage}
\usepackage{extramarks}
\usepackage{fancyhdr}
\usepackage{lastpage}
\usepackage{setspace}
\usepackage{soul}
\usepackage{xspace}

\usepackage[bookmarks=true,pagebackref=true,breaklinks=true,colorlinks,citecolor=gray]{hyperref}
\usepackage{url}

\usepackage{algorithm, algorithmic}
\usepackage{enumerate}
\usepackage[draft]{todonotes} % conflict with CVPR/ICCV format

\newcommand{\subparagraph}{}
\usepackage{titlesec}
\newcolumntype{L}[1]{>{\raggedright\let\newline\\\arraybackslash\hspace{0pt}}m{#1}}
\newcolumntype{C}[1]{>{\centering\let\newline\\\arraybackslash\hspace{0pt}}m{#1}}
\newcolumntype{R}[1]{>{\raggedleft\let\newline\\\arraybackslash\hspace{0pt}}m{#1}}

\newcommand{\sect}[1]{Section~\ref{#1}}

\newcommand{\fig}[1]{Figure~\ref{#1}}
\newcommand{\tbl}[1]{Table~\ref{#1}}

\newcommand{\ignore}[1]{}

\makeatletter
\DeclareRobustCommand\onedot{\futurelet\@let@token\@onedot}
\def\@onedot{\ifx\@let@token.\else.\null\fi\xspace}

\def\eg{e.g\onedot} 
\def\ie{i.e\onedot} 
 
 \def\vs{vs\onedot}
 
\def\etal{et al\onedot}
\makeatother

\definecolor{MyDarkBlue}{rgb}{0,0.08,1}
\definecolor{MyDarkGreen}{rgb}{0.02,0.6,0.02}
\definecolor{MyDarkRed}{rgb}{0.8,0.02,0.02}
\definecolor{MyDarkOrange}{rgb}{0.40,0.2,0.02}
\definecolor{MyPurple}{RGB}{111,0,255}
\definecolor{MyRed}{rgb}{1.0,0.0,0.0}
\definecolor{MyGold}{rgb}{0.75,0.6,0.12}
\definecolor{MyDarkgray}{rgb}{0.66, 0.66, 0.66}

\newcommand{\mypara}[1]{\vspace{3pt}\noindent\textbf{#1}\;\;}

\newcommand{\model}{DensePhysNet\xspace}

\titlespacing\section{0pt}{5pt plus 2pt minus 2pt}{3pt plus 2pt minus 2pt}
\titlespacing\subsection{0pt}{3pt plus 1pt minus 1pt}{2pt plus 1pt minus 1pt}

\begin{document}

% paper title
%\title{Active Self-supervised Learning for Object Physical Property Discovery}
%\title{\model: Learning Dense Physical Object Representations via Multi-Step Interactions} 
\title{\model: Learning Dense Physical Object Representations via Multi-step Dynamic Interactions} 
%for\\Flexible Manipulation 
% You will get a Paper-ID when submitting a pdf file to the conference system
% \author{Author Names Omitted for Anonymous Review. Paper-ID [149]}
\author{Zhenjia Xu$^{1,2}$, Jiajun Wu$^2$, Andy Zeng$^{3,4}$, Joshua B. Tenenbaum$^{2,5}$, Shuran Song$^{3,4,6}$\\
$^1$Shanghai Jiao Tong University $^2$Massachusetts Institute of Technology $^3$Princeton University
\\$^4$Google $^5$MIT Center for Brains, Minds and Machines $^6$Columbia University
}

\maketitle

\begin{abstract}

% In this work, we aim to learn the latent physical properties of objects for manipulation. through dynamic interactions. our goal is to study extent to which we can learn latent physical properties of objects through 

% Intelligent manipulation benefits from the ability to distinguish between object materials and infer their physical properties from sight. Is is possible for robots to acquire this ability without explicit supervision? In this work, we propose \model, a system that actively executes a sequence of dynamic interactions (\eg, sliding and colliding), and uses a deep predictive model over its visual observations to learn dense pixel-wise representations that reflect the physical properties of observed objects.
%We study the problem of learning physical object representations for robot manipulation. 
%Understanding object physics is critical for successful object manipulation: it takes more effort to pick up a metal than a wooden block. The problem is also challenging, because physical object properties can rarely be inferred from appearance cues alone: aluminum has a much lower density than steel, but they look similar. 

We study the problem of learning physical object representations for robot manipulation. 
Understanding object physics is critical for successful object manipulation, but also challenging because physical object properties can rarely be inferred from the object's static appearance. 
In this paper, we propose \model, a system that actively executes a sequence of dynamic interactions (\eg, sliding and colliding), and uses a deep predictive model over its visual observations to learn dense, pixel-wise representations that reflect the physical properties of observed objects.
Our experiments in both simulation and real settings demonstrate that the learned representations carry rich physical information, and can directly be used to decode physical object properties such as friction and mass. The use of dense representation enables \model to generalize well to novel scenes with more objects than in training. With knowledge of object physics, the learned representation also leads to more accurate and efficient manipulation in downstream tasks than the state-of-the-art. Video is available at \url{http://zhenjiaxu.com/DensePhysNet}
%%In this paper, we propose \model, learning dense physical object representations through self-supervised dynamic interactions. \model uses a recurrent encoder to produce a dense, pixel-wise representations from sequences of actions designed to most reveal object physics.

\end{abstract}

\IEEEpeerreviewmaketitle

\section{Introduction}

% motivation 
% Human's ability of inferring intrinsic physical properties from visual data, without explicit supervision. 
Intelligent manipulation benefits from the ability to distinguish between object materials and infer their physical properties from sight. For example, for tabletop object rearrangement, differentiating heavy and light materials enables better planning of manipulation strategies. Is it possible for robots to self-learn these differences without any explicit supervision?

Although considerable research has been devoted to learning object-centric representations that reflect visual features, they rarely account for latent physical attributes such as mass or friction. Unsupervised learning of physical properties is a less explored problem due to three major challenges:
\begin{itemize}[leftmargin=*] % left margin reduces indent (remove if necessary)
    \item Most physical attributes cannot be directly inferred from appearance cues alone in a static environment. For example, while aluminum shares a similar appearance with steel, it is much lighter. %significantly less dense than the other.
    \item Most physical attributes are not salient under static or quasi-static interactions: gently pushing a wooden or a metal block results in only subtle differences in their visible motion, despite their different materials and densities. %metal block can be much heavier. %two drastically different materials. 
    \item Each physical property may only be revealed under specific types of interactions. For example, the sliding distance of an object is determined by both its friction coefficient and mass given its initial momentum; but it is only determined by the object's friction coefficient given its initial velocity. Therefore, without an explicit physics model, the system needs not only to explore different types of interactions, but also to infer and decouple physical properties from multiple action outcomes jointly.
    % a system needs not only to explore different interactions, but also to simultaneously decouple physical properties from each other based on the observed outcomes of the interactions. 
    % the observed outcomes of the interactions %. multiple action outcomes.
%Disentangling these attributes to make them useful for future manipulation tasks is a difficult problem.
\end{itemize}

\begin{figure}[t]
\centering
\includegraphics[width=\linewidth]{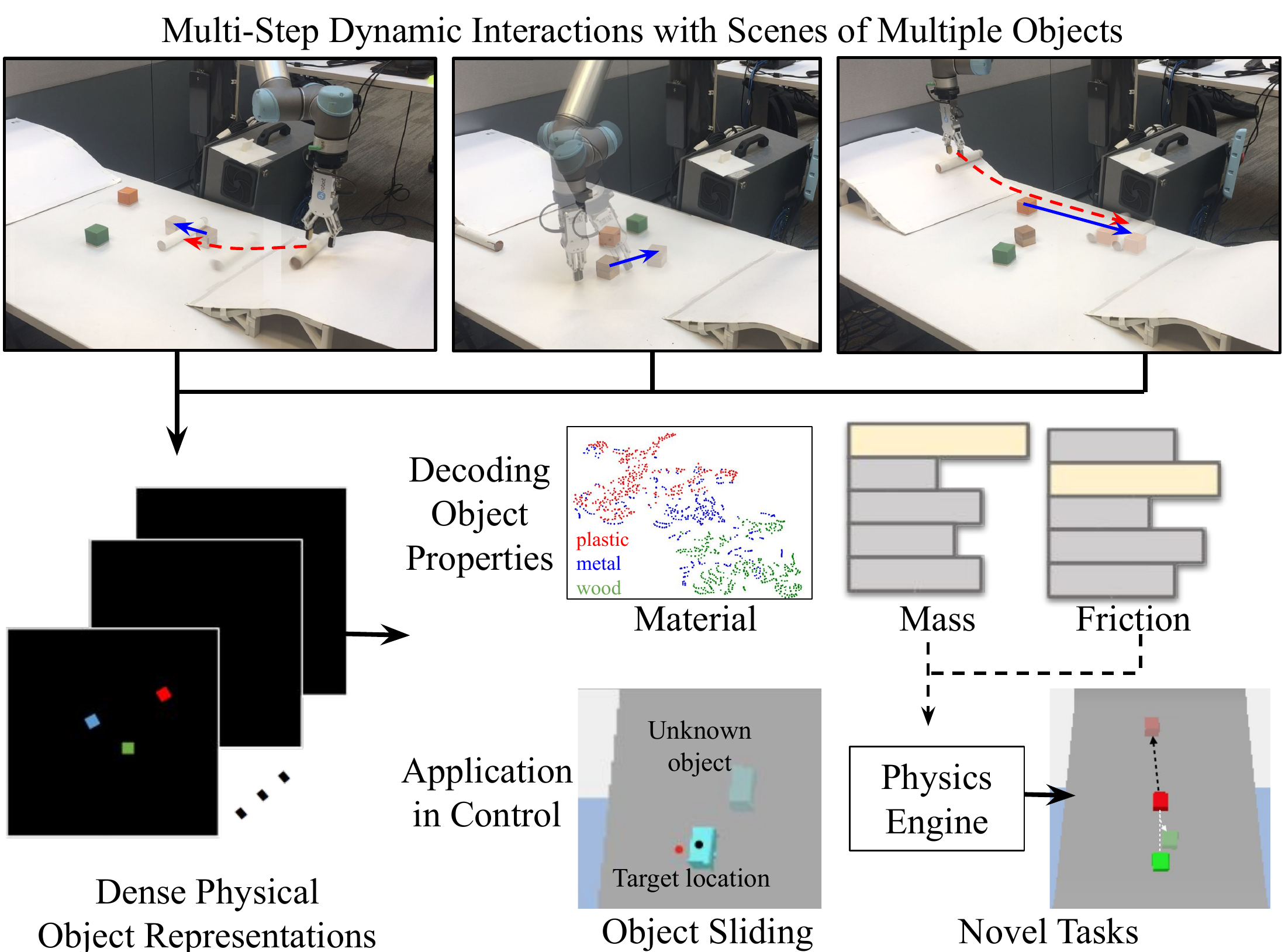}
\vspace{-20pt}
\caption{Our goal is to build a robotic system that learns a dense physical object representation from a few dynamic interactions with objects. The learned representation can then be used to decode object properties such as its material and mass, applied in manipulation tasks such as sliding objects with unknown physics, and combined with a physics engine to tackle novel tasks.
}
\label{fig:teaser}
\vspace{-20pt}
\end{figure}

In this work, we propose to discover and learn the physical properties of objects through visual observations of \textbf{multi-step, self-supervised, dynamic interactions}. Our system, \model, actively executes a sequence of dynamic interactions (\eg, sliding and colliding), and uses its visual observations to learn physical object properties without any explicit supervision (\fig{fig:teaser}).

\model takes the current scene and action as input and predicts how objects move after the interaction, represented as pixel-wise optical flow. 
%By predicting the future state of objects under a variety of dynamic interactions, \model learns to discover and model physical object properties. 
By learning to predict future object states conditioned on different interactions,  \model acquires an implicit understanding of their physical properties and how they influence observed motions. We use a recurrent structure to aggregate information from multiple interactions, so that \model can better infer and encode objects' physical properties over time. We also design \model in a modularized way, so that the learned physical representations can be disentangled from visual representations, which encode objects' visual appearance and actions. This unique design enables it to generalize to new tasks that involve different types of interactions. Because \model produces a pixel-wise dense representation, instead of a single feature vector for the entire scene as in many prior methods~\cite{pinto2017learning,Li2018Push,Agrawal2016Learning}, it also generalizes to complex scenes that contain more objects than training scenes. %, even when trained only on single object scenes. 
%\jw{do we have results on generalizing to more complex scenes?} \shuran{it can work for three object?}
% result 

The main contribution of our paper is to demonstrate that deep predictive models over visual observations of multiple dynamic interactions can enable an agent to learn the latent physical attributes of manipulated objects. We execute a series of experiments in both simulation and real settings to evaluate our approach qualitatively and quantitatively. The results show that \model is more effective at learning physical object properties than other representation learning methods, and that the learned representations can be leveraged to improve the performance of downstream control tasks such as planar sliding. %  by suggesting more accurate actions policies, both in simulation and on a real robot. 
We also show that when combined with a physics engine, \model is capable of leveraging the decoded physical object properties to execute novel control tasks. 

\section{Related Work}

Modeling object physics is a long-standing problem in both robotics and artificial intelligence. Dating back to the 1980s, Atkeson~\etal~\cite{atkeson1986estimation} estimated the mass and moments of inertia of a grasped object, using measurements from a wrist force-torque sensor. Yu~\etal~\cite{yu2005estimation} tackled the same problem by pushing objects using two fingertips equipped with force-torque sensors, and recording the fingertip forces, velocities, and accelerations. Recently, researchers have explored learning to estimate physical properties from their appearance and motion, either in combination with physical simulators~\cite{Wu2015Galileo,Wu2017Learning} or via end-to-end deep learning~\cite{Ehrhardt2017Taking,Fragkiadaki2016Learning,ye2018interpretable,Watters2017Visual}. These models build upon an explicit physical model, \ie, a model parameterized by physical properties such as mass and force. This enables generalization to new scenarios, but also limits their practical usage: annotations on physical parameters in real-world applications are expensive and challenging to obtain. 

An alternative line of work is to learn object representations without explicit modeling of physical properties, but in a `self-supervised' way through robot interactions. Byravan and Fox proposed to use deep networks to approximate rigid object motion~\cite{byravan2017se3}. Pinto~\etal~\cite{Pinto2016Curious} proposed to build a `curious' robot that interacts with objects to learn representations for visual recognition. Denil~\etal~\cite{Denil2017Learning} used reinforcement learning to build object models via physical experiments. Recently, Zheng~\etal~\cite{Zheng2018Unsupervised} suggested explicit physical properties can be decoded from the latent representations learned through interactions. A few follow-ups have extended these models for planning and control, including poking~\cite{Agrawal2016Learning}, pushing via transfer learning~\cite{pinto2017learning}, and visual predictive learning~\cite{Finn2017Deep}. There have also been several papers on getting better policies via modeling environment dynamics~\cite{yu2017preparing, zhou2018environment}. Such progress is impressive, though the focus of these papers is still on \textit{visual} representation learning. Without considering the underlying physical processes, they fall short to generalizing to novel objects and tasks.

The work closest to ours is Push-Net, proposed by Li~\etal~\cite{Li2018Push}, which uses a multi-step model to learn physical object properties for planar pushing. While Push-Net focuses on scenes with a single object and encodes the entire image into a single latent representation, we instead learn dense (pixel-wise) representations from interactions -- which enables our model to generalize not only to novel objects, but also to novel scenes with multiple objects. %Since we are able to decode explicit physical properties form the latent representation, we are also able to use these explicit physical properties to support other tasks other than planar pushing. 
Furthermore, while Push-Net only considers planar quasi-static pushing, we consider two types of dynamic interactions: planar sliding and collision. Our experiments demonstrate that using multiple types of dynamic interactions is key to fully revealing latent physical properties.
%\jw{i removed the sentence on task generalization. their model, when paired with some supervision on physical properties, can also be applied to new tasks.}
%Recently, people have used neural nets to approximate the pair -wise interaction between objects for simulating forward dynamics~\cite{Battaglia2016Interaction,Chang2017compositional}. 
%Unsupervised  Learning  of  Latent  Physical  Properties  Using Perception-Prediction  Networks:
%ShapeStack

%\subsection{Dense Object Representations}

We are, of course, not the first to learn dense representations on visual data. Most prior work on this topic revolve around learning correspondences across views in 2D~\cite{choy2016universal,schmidt2017self} and 3D~\cite{Zeng20173DMatch,thewlis2017unsupervised,shotton2013scene,brachmann2014learning}. Florence~\etal~\cite{florence2018dense} proposed dense object nets, learning dense descriptors by multi-view reconstruction and applying the descriptors to manipulation tasks. While their paper primarily focuses on learning object representations that reflect visual appearance, we learn object representations that reflect physical properties and show that they are useful for manipulation tasks that require physical knowledge.

\section{Method}

\begin{figure}[!t]
\centering
\includegraphics[width=\linewidth]{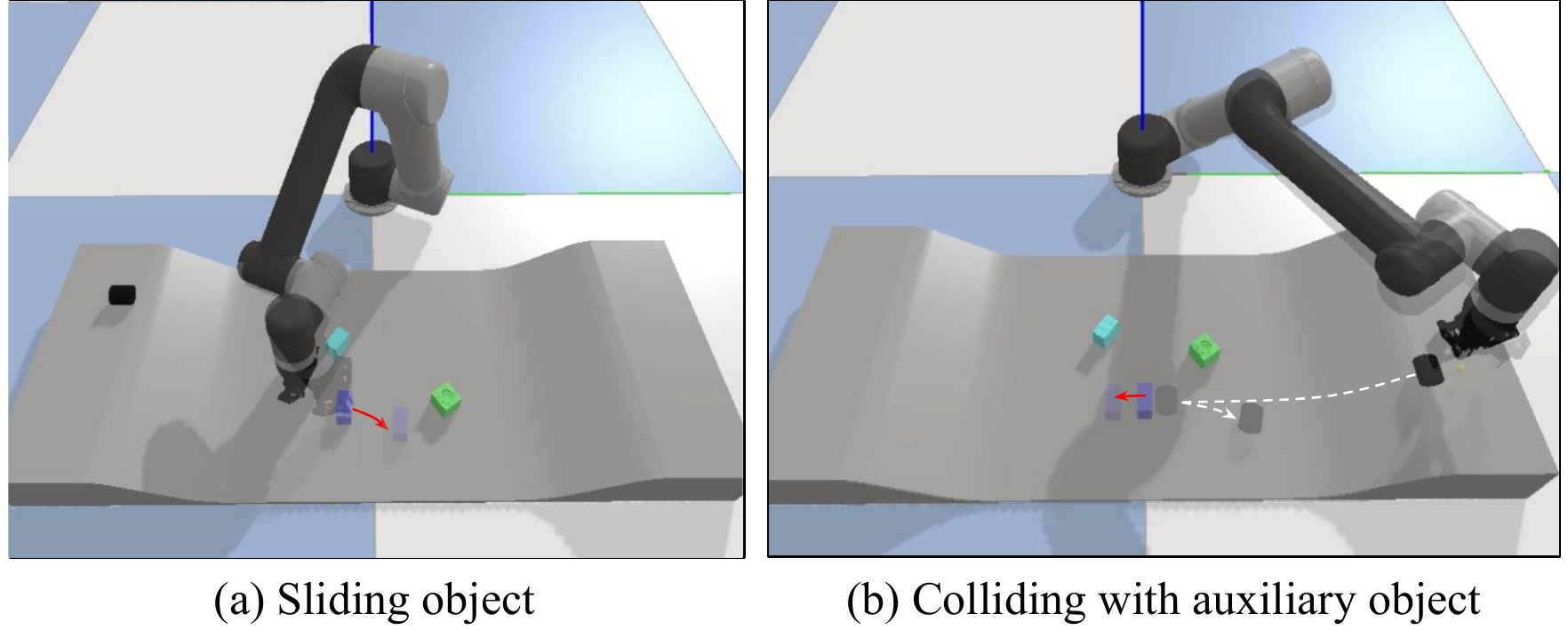}
\vspace{-20pt}
\caption{\textbf{Dynamic Interactions}. We design two types of dynamic interactions  (sliding and collision) to reveal physical object properties. To slide an object, the robot approaches it from an angle and executes a push with a high speed, such that the object can slide after the push. The robot releases an auxiliary object on the ramp to make collisions.
}
\vspace{-15pt}
\label{fig:actionspace}
\end{figure}

The goal of \model is to learn latent representations that encode object-centric physical properties (\eg, mass, friction) through self-supervision. To this end, we train a deep predictive model of depth images on a large dataset of observed dynamic robotic interactions. 
The idea is that in order for \model to accurately predict the future states of objects conditioned on different interactions, it needs to acquire an implicit understanding of objects' physical properties and how they influence observed motion.
The setting consists of a collection of objects on an inclined ramp laid in front of the robot (\fig{fig:actionspace}). When interacting with objects, the robot captures a depth image $I_t$ of the state at time $t$, executes an action $a_t$, then captures another depth image $I_{t+1}$ at the next time step. We model \model as a neural network that takes as input the visual observation $I_t$ and the executed action $a_t$, and outputs a prediction of the next state observation $I_{t+1}$ in the form of optical flow $O_{t,t+1}$.  

%To do that, the system needs to choose an action $a_t$ to interact with the scene at time $t$. Denote $I_t$ and $I_{t+1}$ as the depth map of the scene before and after the interaction, respectively. The network takes in the action and current observation as input $\{a_t,I_t\}$, and output the prediction of future state in the form of optical flow $O_{t,t+1}$ between $I_t$ and $I_{t+1}$. The idea is that, in order for \model to accurately predict the future state of the object, it needs to learn about objects' physical properties that influence its motion under different interactions. 

\begin{figure*}[!t]
\centering
\includegraphics[width=\linewidth]{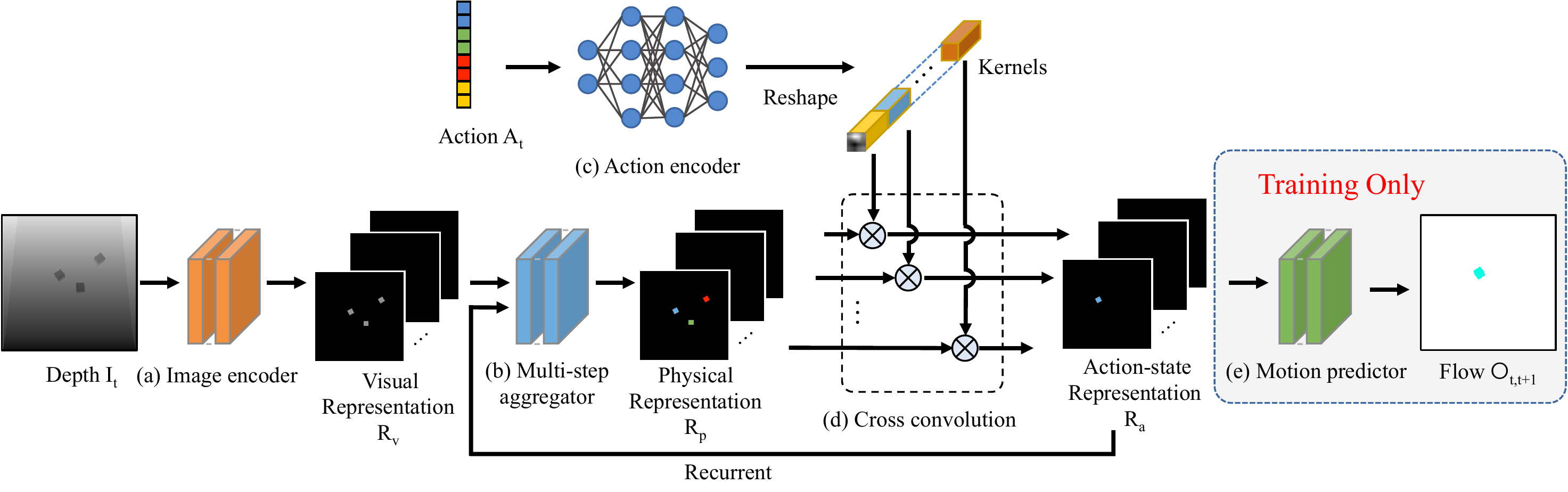}
\vspace{-25pt}
\caption{\textbf{\model}. \model takes in the current state (depth image $I_t$) and interaction  $A_t$ as input and predicts the change of objects' state after the interaction. The change of objects' state is represented as pixel-wise optical flow $O_{t,t+1}$. 
The network consists of five modules: (a) an image encoder, (b) a multi-step information aggregator, (c) an action encoder, (d) a cross convolutional layer, and (e) a motion predictor. These five modules work jointly to learn three object representations: visual representations $R_v$ that encode visual signals of the object, physical representations $R_p$ that encode physical object properties, and action-state representations $R_a$ that encode objects' states after interaction. 
%\jw{label depth image $I_t$, Action $A_t$, Visual representation $R_v$, optical flow $O_{t,t+1}$ , action kernels $XXXX$ etc. check text carefully} \jw{relabel image encoder as a, aggregator as b, action encoder as c}
}
\vspace{-15pt}
\label{fig:model}
\end{figure*}

\subsection{Dynamic Interactions}
The robotic agent executes two types of dynamic interactions to accentuate the physical properties of objects: sliding and collision. The vast majority of prior methods in representation learning~\cite{Li2018Push,Finn2017Deep,pinto2017learning,Agrawal2016Learning} use static or quasi-static manipulations like pushing, grasping, or poking, where it is challenging to observe latent physical object properties from object motion. This is because in (quasi-)static manipulation, changes in object movements are largely influenced by the actions and dynamics of the manipulator, less so from the object itself. For example, during quasi-static pushing, the object is assumed to move only with the end effector and to stop when the end effector stops. In these scenarios, it is naturally more difficult to observe the subtleties in motion due to the physical properties of the object.

Dynamic manipulations, in contrast, can lead to object motions beyond contact with the manipulator that are more likely to reveal its physical properties. For example, when objects are pushed with a high speed, the forces of acceleration can cause objects to slide by themselves for a distance. The differences in their motion after contact, \eg, distance traveled, can serve as a visual cue for differences in surface friction. These cues are relatively more salient than what can be observed in quasi-static manipulations. However, the distance traveled by an object after sliding alone, assuming known initial velocity, can only help derive the object's friction. Therefore, we need other types of dynamic manipulations such as collisions to reveal and distinguish other physical properties.

% When objects are pushed with high speeds, the forces of acceleration can cause the object to slide by itself for a distance.  during the course of which its motion (\eg distance traveled) is largely influenced by its own surface friction (less so from mass). Likewise, when an object collides with another, the motion trajectories of both objects are influenced by both their mass and friction.
% Using a combination of both sliding and collision can help to disentangle these properties.
% The differences in motion trajectories from dynamic interactions are more visually salient. Hence, we use both sliding and collision dynamic interaction and combine both sliding and collision to eliminate the ambiguity.

\mypara{Sliding.} The sliding action is parameterized by direction $\theta$ and velocity $v$ with respect to the robot. To slide an object, the robot approaches it from $\theta$ and executes a push with a high speed such that the object can slide after the push.  
To achieve high pushing velocities without exceeding the physical force-torque safety limits of the real-world robot, we constrain the motion planning so that joints closer to the base of the arm (\eg, elbow, shoulder) move slower than joints closer to the end effector (\eg, wrist). The direction $\theta$ and velocity $v$ are quantized as discrete variables and we use one-hot vectors to encode them. There are sixteen possible directions and four possible speeds. The sixteen directions are uniformly distributed on $[0, 2\pi]$. Four robot's speeds are $\{0.96,1.28,1.44,1.6\}$rps, representing the robot's joint rotation speed.
%\shuran{@andy add some explanation here. it is in fact not the wrist, it is the second to last joint}
%$\{0.03, 0.04, 0.045, 0.05\}$ \jw{add unit}

\mypara{Collision.}
For collisions, we set two inclined ramps, one on each side of the workspace. The robot grasps an auxiliary cylinder and places it on the top of one of the ramps. The cylinder then rolls down the ramp and collides with an object in the middle. The only parameters of the collision action are the cylinder's starting position $[d \times X_s, y, H_s]$, where $d\in\{-1, 1\}$ indicates whether the cylinder is rolling from the left or right ramp, $X_s$ is the distance from the ramp to the center of workspace, $y$ is the same as the $y$-coordinate of the target object, and $H_s$ is the height of the ramp. The auxiliary cylinder is fixed with radius $2$cm, height $4$cm, and weight $0.4$kg. 

\mypara{Interaction policy.}
We use a balanced-random interaction policy to ensure interaction diversity and to encourage a full exploration of each action. Specifically, for each step, we choose one type of action to execute via $P(x_{i, j}) = (1/2)^{a_{i, j}}/\sum_{n=1}^N\sum_{m=1}^M (1/2)^{a_{n, m}}$, where $P(x_{i, j})$ is the probability to apply action $i$ to object $j$, $N$ is the number of objects in the scene, $M$ is the size of action space, and $a_{i,j}$ is the number of action $i$ applied to object $j$ up to now. We randomly sample the parameters for the chosen action, while enforcing the constraint that the object needs to stay inside the workspace.  

\subsection{\model}

We design \model based on three key insights. First, it needs to be modularized in a way that the learned physical representations are disentangled from the representations that encode information about object visual appearance and actions. This is critical for applying the learned physical representations to new tasks, objects, and action types. Second, \model has a recurrent structure, so that it can aggregate information from multiple interactions to better infer physical properties. Third, our model produces a dense pixel-wise representation, instead of encoding the entire scene into one single latent representation as in previous papers~\cite{Li2018Push}. This enables handling complex scenes with multiple objects.  

%In the following sections, we provide details on the network architecture and the action space. 

% Our model is trained by predicting future motions with optical flow as supervision. 
% During testing, our model only needs a sequence of images and actions, without any extra supervision.
% \fig{fig:model} shows an overview of our model.

% The robot interacts with the object for several times and get a sequence of image and action data $D = {(I_1, A_1), (I_2, A_2) \cdots (I_n, A_n)}$.
% Here $A_k$ is the action in step $k$. $I_k$ is the depth image after $A_{k-1}$ and it is also the state before $A_k$.

\model (\fig{fig:model}) consists of five modules: an image encoder, a multi-step information aggregator, an action encoder, a cross convolutional layer, and a motion predictor. These five modules work jointly to learn three object representations: visual representations $R_v$ that encode visual signals of the object, physical representations $R_p$ that encode physical object properties, and action-state representations $R_a$ that encode objects' states after interaction. 

%It has a recurrent structure to handle the sequence of data. In each step $t$, it takes the current image $I_t$ and action $a_t$ as input, update the representations, and predict the optical flow $O_{t,t+1}$ after this action.
%\shuran{I have hard time to understand the network overview below.@jiajun could you help?}
Each iteration of learning works as follows. First, given a depth image $I_t$, the image encoder (\fig{fig:model}a) extracts visual signals from the image and outputs the visual presentation $R_v$. Then, the information aggregator (\fig{fig:model}b) learns to integrate the visual representation $R_v$ with the object representation after the last interaction $R_a$ to extract the physical representation $R_p$. Intuitively, after interactions, objects with different physical properties will end up in different positions and poses, and such signals are now available from the visual input and therefore should lie within the visual representation $R_v$. The goal of the information aggregator is thus to distill physical knowledge $R_p$ by analyzing $R_v$ and $R_a$ jointly.  
%There are three kinds of representations in the model: visual representation ($VR$), physical representation ($PR$), and action-state representation ($AR$).
%Intuitively, we can infer some information of physical property after observing the tuple ($I_{t-1}$, $A_{t-1}$, $I_t$) %Here the information of $I_{t-1}$, $A_{t-1}$ and several previous understanding to this object will be encoded in $AR_{t-1}$. %Then, the multi-step aggregator (\fig{fig:model}c) combines $AR_{t-1}$ and $VR_t$ to get a better understanding of this object, and this this will be encoded in physical representation ($PR_t$).

In parallel, the action encoder (\fig{fig:model}c) encodes action $a_t$ into convolution kernels. The cross convolutional layer (\fig{fig:model}d) then applies the encoded action kernels on the physical representation $R_p$ to produce the effect of the action on the objects. Here, the cross convolutional layer can be seen as a learned, latent physical simulator, learning physics to approximate the effect of actions. As shown in Xue~\etal~\cite{Xue2016Visual}, this cross convolutional layer better integrates the action and the physical representations compare to tensor concatenation. %Our model then performs convolutions (\fig{fig:model}e) on the physical representation ($PR_t$) using separate kernels obtained from an action decoder (\fig{fig:model}a) and get the action-state representation ($AR_t$) in this step .
It outputs the action-state representation $R_a$. %Compared with the physical representation $R_p$, the action-state representation $R_a$) is endowed the in information of action ($a_t$).

Finally, the action-state representation $R_a$ is fed into the motion predictor (\fig{fig:model}e) to predict the optical flow $O_{t,t+1}$ across two images $I_t$ and $I_{t+1}$. This is the only supervisory signal we have during training; during testing, the motion predictor is no longer needed. %predict the motion optical flow $O_t$ by using an motion predictor (\fig{fig:model}e).

%\subsection{Implementation Details}

%Following sections provides mode details on each of the network module. 

\mypara{Network architecture.}
Here we provide more details on each network module.  The image encoder takes as input the depth image with a size of $160\times 160$, and outputs a $32$-channel feature map. It has one $11\times11$, one $5\times5$, and two $3\times3$ convolutional layers. There are batch normalization and ReLU layers between adjacent convolutional layers.

%\mypara{Action encoder.}
The action encoder uses seven fully connected layers with $64, 128, 256, 256, 256, 512$, and $800$ hidden units, respectively. It takes in the action vector (a $37$-dim vector) an input and outputs $32$ action kernels, each with a size of $5\times5$.

%\mypara{Multi-step aggregator.}
The multi-step aggregator takes the visual representation $R_v$ and the action-state representation $R_a$ in the last step as input (both have a size of $32\times 160\times 160$), and combines them with a gate $C$ via  $C = S(g(R_v)) f(R_v) + \left[1 - S(g(R_v))\right] R_a$, where $S(\cdot)$ is the sigmoid function and $f(\cdot)$ and $g(\cdot)$ are a $1\times1$ convolution layer. The gate acts as a filter to block or pass on information based on the calculated weight. Then, fifty residual blocks~\cite{He2016Deep} are applied to the combination $C$ to get the physical representation.

%\mypara{Cross convolution.}
% \shuran{again why use cross conv? what's the property of crossconv make it good to combining action and state? } \jw{agreed.}
%The action can be considered as a transformation to object position. Compared to tensor concatenation, the cross convolutional layer~\cite{Xue2016Visual} provides a better combination to the action and the physical representation. 
The cross convolutional layer applies the convolutional kernels from the action encoder to the physical representation $R_p$ from the multi-step aggregator to produce the action-state representation $R_a$. Here, the convolutions are carried out in a channel-wise manner, \ie, each of the $32$ layers in the physical representation is convolved with one of the $32$ kernels.

%\mypara{Motion predictor.}
The motion predictor takes the action-state representation $R_a$ as input and outputs a pixel-wise optical flow map. This module consists of four $3\times3$ convolution layers with $32, 32, 32$, and $2$ channels, respectively. In between, there are batch normalization and ReLU layers.

\mypara{Self-supervised training.}
We use Mean Square Error (MSE) between the predicted optical flow and ground truth as the training loss. Because optical flow can be automatically computed based on visual observations (\ie, images taken before and after the interaction), the whole training process of \model can be fully self-supervised without any human annotations. We implement our model in PyTorch~\cite{Paszke2017Automatic}. Optimization is carried out using ADAM~\cite{Kingma2015Adam} with $\beta_1 = 0.9$ and $\beta_2 = 0.95$. We use an initial learning rate of $10^{-3}$ and a learning rate decay of $0.9$ after each epoch. The model is trained for $40$ epochs with a mini-batch size of $32$.

\section{Experiments}
\label{sec:exp}
We now evaluate the learned physical representation $R_p$. The goal of the experiments is to understand, through both qualitative analysis and quantitative evaluation, whether the learned physical representations encode information about physical object properties; if so, how accurate the encoding is, and how useful the representation is for object manipulation. %We answer these questions through both qualitative analysis and quantitative evaluation. 

\subsection{Decoding Object Material}
\label{sec:material}

We first analyze whether the learned physical representation can be used to distinguish objects of different materials. 

\mypara{Baselines.} We compare our model for learning physical representations $R_p$ with two baselines to understand the role of active interaction.
\begin{itemize}[leftmargin=*]
    \item Visual representations: we first compare the physical representations $R_p$ with the learned visual representations $R_v$ extracted from our model (see \fig{fig:model}). The learned visual representations are computed directly from static images, without the knowledge of object motion or actions. 
    \item Our model with sliding: we also compare with our model trained with only sliding, but not collisions, to validate the importance of having multiple types of interactions.
\end{itemize}

\mypara{Setup.} For experiments in simulation, we use three visually indistinguishable objects, made of one of the three materials: 
\begin{itemize}
    \item Plastic with mass $m\in[0.11, 0.14]$kg and friction coefficient $\mu \in[0.4, 0.6]$.
    \item Wood with mass $m\in[0.11, 0.14]$kg and friction coefficient $\mu \in[0.8, 1.0]$.
    \item Metal with mass $m\in[0.17, 0.20]$kg and friction coefficient $\mu \in[0.4, 0.6]$.
\end{itemize}

Each object's physical properties are uniformly sampled according to its material.
Hence, objects of different materials have distinct physical properties, and those made of the same material have similar, but not identical, properties. We use $8{,}000$ sequences for training. We use PyBullet for all our experiments in simulation~\cite{Coumans2010Bullet}.
For each trial, the robot interacts with the objects $19$ times (in test mode without optical flow supervision). We use background subtraction to compute the silhouette of each object; we then extract pixel-wise features for each silhouette as object representations. 

\begin{figure}[!t]
%\vspace{-3mm}
\centering
\includegraphics[width=\linewidth]{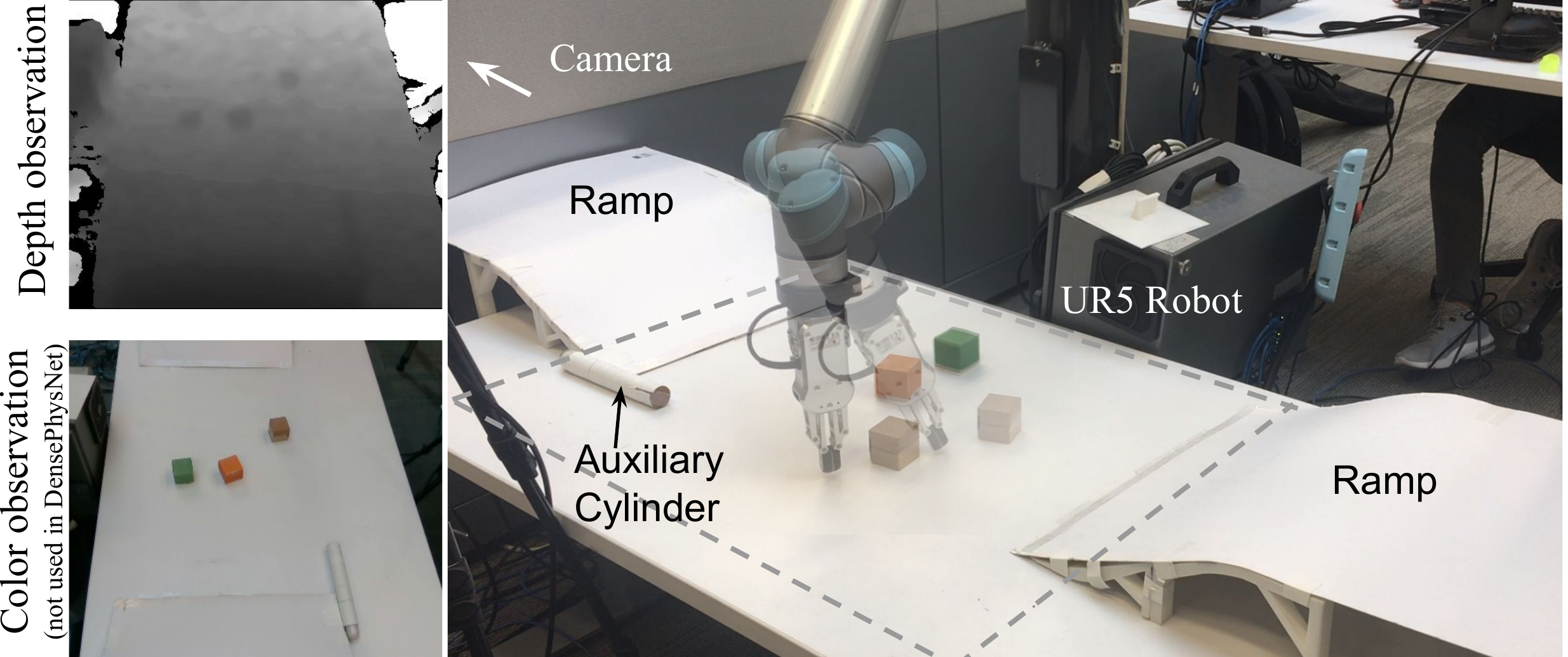}
\vspace{-15pt}
\caption{\textbf{Real-world experiment Setup.} The real-world settings includes UR5 robot arm with an RG2 gripper. A flat table with two incline ramp on both side for collision. We capture RGB-D images using a calibrated Intel RealSense D415, mounted on a fixed tripod overlooking the table from the side.
}
\vspace{-5pt}
\label{fig:real_setup}
\end{figure}

\begin{figure}[!t]
\centering
\includegraphics[width=\linewidth]{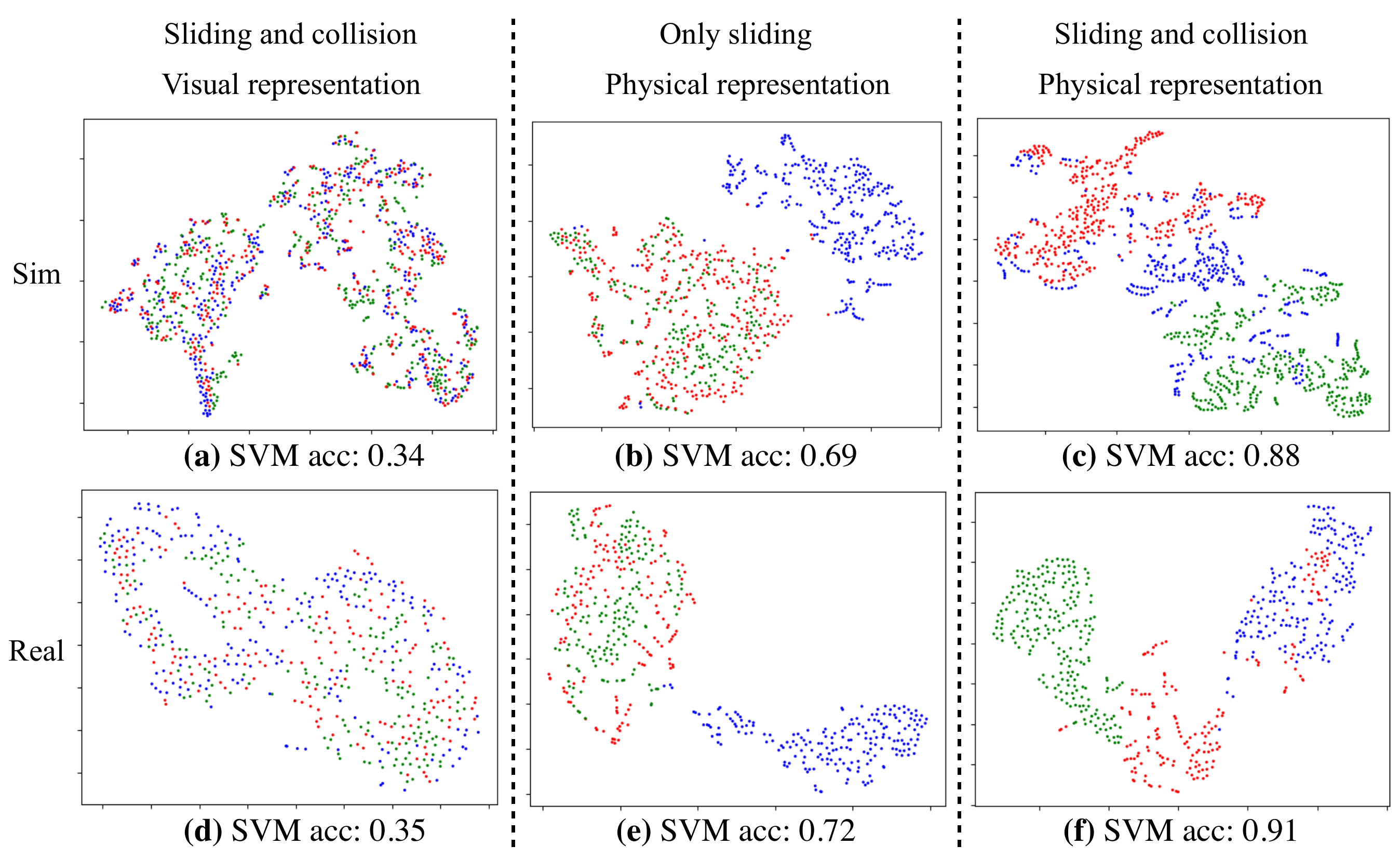}
\vspace{-15pt}
\caption{\textbf{t-SNE embeddings of learned features.} The first row shows results in simulation and the second shows results of real-world experiments. We also compare different representations and actions. For each result, we train a 3-way SVM for material classification; its accuracy is presented under the embedding visualization. For simulation results, the visual representation (a) is not informative in distinguishing object material. The representation learned from sliding alone (b) is only informative in distinguishing the difference in friction (wood \vs others), but has trouble telling the difference in mass (plastic \vs metal). Our \model (c), designed to learn physical object representations, learns to cluster objects based on their material. Also, in real-world experiments, only the physical representation learned from both interactions (f) can distinguish different materials. 
% \jw{I'd say remove the table and just have 6 figures. For each figure, label the actions and representation type on top, and put classification score somewhere on the figure.}
% \shuran{ move the SVM score next to figure title? e.g. (a) SVM acc: 0.34. Now it is hard to understand what's the meaning of the numbers.}
} %\shuran{push only instead of only push?}}
\label{fig:visualization_result}
\vspace{-15pt}
\end{figure}

We also evaluate \model in real-world settings, where we use a UR5 robot arm with an RG2 gripper. \fig{fig:real_setup} shows the setup. We capture RGB-D images using a calibrated Intel RealSense D415, mounted on a fixed tripod overlooking the table of objects from the side. The camera is localized using an automatic calibration procedure \cite{zeng2018learning}.
The pushing and collision primitives are open-loop, with robot arm motion planning executed using IK solvers~\cite{diankov_thesis}. %To achieve high pushing velocities without exceeding the physical force-torque safety limits of the UR5, we constrain the motion planning so that joints closer to the base of the arm (\eg elbow, shoulder) move slower than joints closer to the end effector (\eg wrist).
In real-word experiments, we use the model pre-trained in simulation. In total, we test on $20$ sequences for three objects of different materials. Each sequence contains fifteen interaction steps. We consider two kinds of interactions policies: (1) sliding only; (2) both sliding and collision. 
For each sequence, we ensure the object is always within the workspace. In total we collect $10$ sequences for each interaction policy.

% There are also some failure cases in real-world experiments. \fig{fig:failure_case} demonstrates two typical failure cases. We will ignore these sequences in the following experiments.
%\input{figText/failure_case.tex}

\begin{figure}[!t]
\centering
\includegraphics[width=\linewidth]{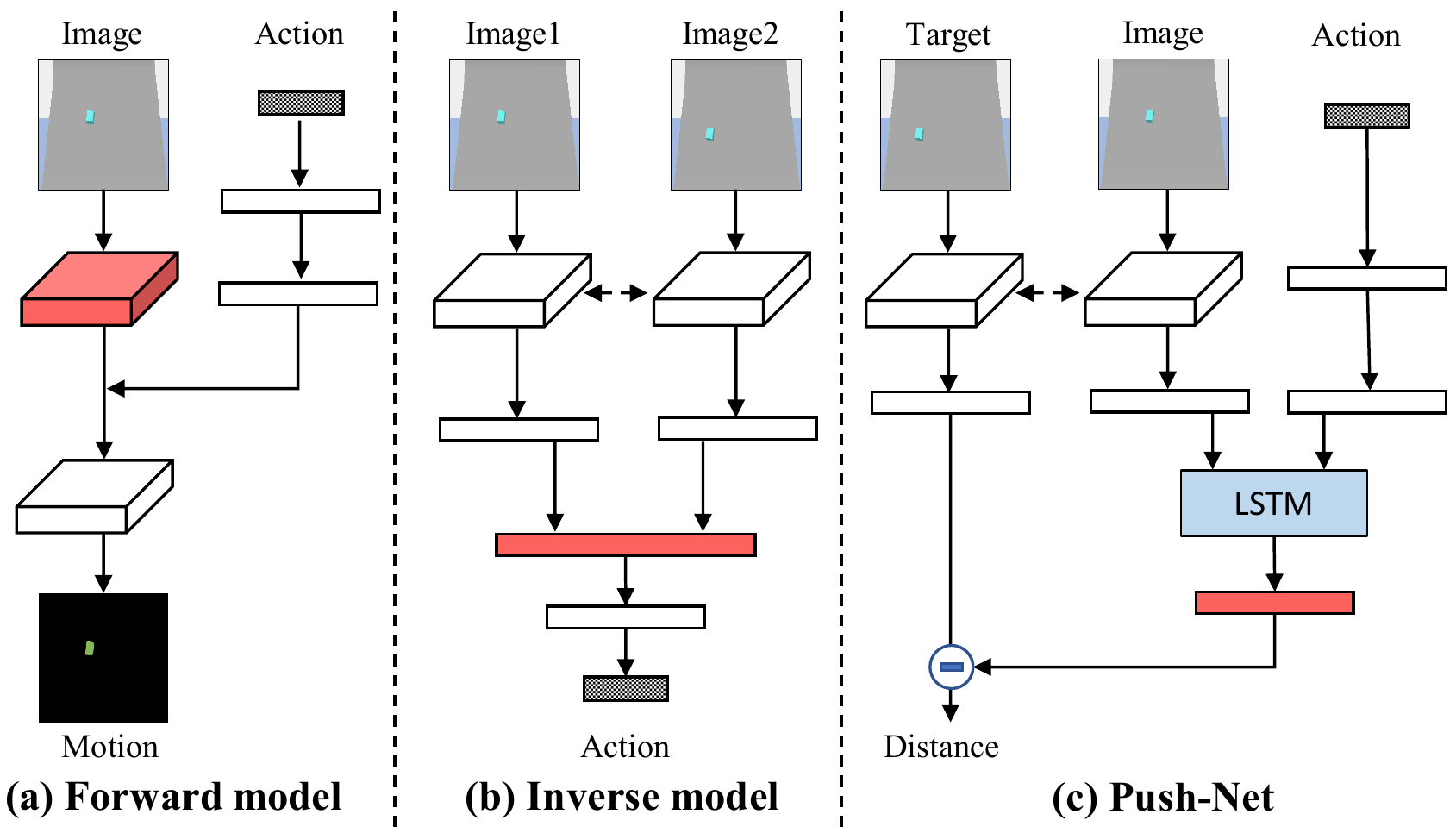}
\vspace{-20pt}
\caption{\textbf{Baseline models.} Agrawal~\etal~\cite{Agrawal2016Learning} proposed a forward and an inverse model, both designed to handle a single-step interaction. The forward model (a) takes the current frame and the action as input, and predict the motion of the object. The inverse model (b) takes the frames before and after the action as input and predicts the action parameters. Push-Net~\cite{Li2018Push} (c) uses an LSTM to capture the history of interactions.
%\jw{we can remove this figure or at least greatly simplify it, to save space for Fig. 4}\shuran{can we simply the figure into a 1x3 grid with three baselines? each of the model with minimal blocks. }
}
\vspace{-15pt}
\label{fig:application_baseline}
\end{figure}

\mypara{Results.} We randomly sample $1{,}000$ pixels within object silhouettes from $100$ test sequences. We then embed the corresponding $1{,}000$ learned feature vectors using 2D t-SNE embeddings~\cite{maaten2008visualizing}. The first row of \fig{fig:visualization_result} shows the results, where colors represent different materials. Our model, designed to learn physical object representations, learns to cluster objects based on their material. The representations learned from sliding alone is only informative in distinguishing the difference in friction (wood \vs others), but has trouble differentiating different mass (plastic \vs metal). The visual representation is not informative in distinguishing object material.

We further verify each representation's discriminative power by training a 3-way linear-SVM on $1{,}000$ feature vectors (pixels) randomly sampled from $100$ training sequences, and testing its material classification accuracy on $100$ feature vectors (pixels) randomly sampled from $100$ test sequences. In simulation, the representation learned by \model achieves a $88\%$ accuracy on material classification, while the model trained only on sliding achieves an accuracy of $69\%$. The visual representation achieves an accuracy of $34\%$, close to random guess ($33.3\%$). In the real-world experiment, \model achieves a $91\%$ accuracy when using both kinds of interactions. If the model only uses sliding, the accuracy is $72\%$. The visual representation achieves an accuracy of $35\%$.

\subsection{Decoding Physical Object Properties}
\label{sec:decode}
In this experiment, we examine how accurately we can decode physical properties from the learned latent representations. 

\mypara{Baselines.} We compare with the following baselines.
\begin{itemize}[leftmargin=*]
    % \item Our model with sliding or collisions: the performance of our model trained with only sliding or collisions, but not both.
    \item Our model with different interaction types.
    \item Single-step forward model~\cite{Agrawal2016Learning}: We also compare our approach with the the model proposed by Agrawal~\etal~\cite{Agrawal2016Learning}. Their model consists of a forward and an inverse model, both designed to handle a single-step interaction instead of aggregating the information across multiple interactions. The forward model takes the current frame and the action as input, and predict the motion of the object for one step (\fig{fig:application_baseline}a).
    \item Single-step inverse model~\cite{Agrawal2016Learning}: The inverse model takes the frames before and after the action as input, and predict the action parameters (\fig{fig:application_baseline}b). 
    \item Push-Net~\cite{Li2018Push}: For each step, the model takes the current mask $M_t$, the action and a target mask $M_{t+1}$ as input, and predict the distance between the underlying states of the target $M_{t+1}$ and the mask generated as a result of applying the action to $M_t$. The original Push-Net also predicts the object's center of mass. For a fair comparison, we omit this part since it requires additional annotation (\fig{fig:application_baseline}c).
\end{itemize}

\mypara{Setup.} To this end, we train a linear classifier to decode physical properties from the latent representations on an annotated dataset, and test it on a set of novel objects of different shapes. During testing, the robot interacts with the objects to update the latent representations, but no optical flow supervision is used. We conduct this experiment in simulation, where ground truth physical properties are available for training. We only evaluate on scenes of a single object, as both the forward and the inverse models in Agrawal~\etal~\cite{Agrawal2016Learning} only handle single-object scenarios and predict one feature vector for the whole scene. The shape of each object is randomly sampled from ShapeNet~\cite{Chang2015Shapenet}, where training and testing objects are of different shapes. Each object's physical properties (friction and mass) are uniformly randomly sampled from $30$ discrete values: $\mu\in\{0.4, 0.41, 0.42, \dots, 0.7\}$ and $m\in\{0.11, 0.113, 0.116, \dots, 0.2\}$kg. There are $8{,}000$ sequences for training and $2{,}000$ for testing. The robot has $9$ interaction steps with the object using three types of policies: sliding only, collision only, and both sliding and collision. We extract the pixel-wise features of the object from the last step using its bounding box (automatically computed via background subtraction), and flatten it into one vector. %and up-sample it to 32$\times$32 with bilinear interpolation. Then we flatten it to vector.
For the forward and the inverse models, we directly take its hidden layer features (marked red in \fig{fig:application_baseline}) as the features of the object.

%\jw{again, repeating ourselves} We train both of the friction and mass decoder as an N-way classifier ($N=30$), which takes in the processed representation as input and predict the probability for each category. 

For evaluation, we train both the friction decoder and the mass decoder as a $30$-way linear classifier using the cross-entropy loss. The classifiers are trained on $1{,}000$ of the $8{,}000$ training sequences and evaluate it on all $2{,}000$ test sequences. We then calculate the weighted distance error for each piece of data as the evaluation metric: $D_\text{err} = \sum_{i=1}^{30} p_i \times |i - y|$, where $p_i$ is the probability of category $i$ predicted by our model and $y$ is the ground truth. We use this metric for many of the following experiments.

% \begin{table}[t]
% \centering\small
% \setlength{\tabcolsep}{3pt}
% \caption{Quantitative results on physical property decoding.}
% \vspace{-5pt}
% \begin{tabular}{lcccccc}
%     \toprule
%     & \multicolumn{2}{c}{Sliding} & \multicolumn{2}{c}{Collision} & \multicolumn{2}{c}{Sliding \& Collision} \\
%     \cmidrule(lr){2-3}\cmidrule(lr){4-5}\cmidrule(lr){6-7}
%     & Friction & Mass & Friction & Mass & Friction & Mass \\
%     \midrule
%     Forward  \cite{Agrawal2016Learning} &10.00&10.00&10.00&10.00&10.00&10.00\\
%     Inverse  \cite{Agrawal2016Learning} &6.41&9.88&8.59&7.96&6.36&6.87\\
%     Ours &4.07&9.23&6.58&7.03&\textbf{3.81}&\textbf{4.24}\\
%     \bottomrule
% \end{tabular}
% \label{tbl:regression_table}
% \vspace{-2mm}
% \end{table}

\begin{table}[t]
\centering\small
\setlength{\tabcolsep}{3pt}
\caption{Physical property decoding error $D_\text{err}$ with different models}
\vspace{-5pt}
\begin{tabular}{lcccc}
    \toprule
    & Forward \cite{Agrawal2016Learning} &
    Inverse \cite{Agrawal2016Learning} &
    Push-Net \cite{Li2018Push} &
    Ours\\
    \midrule
    friction & 10.00 & 6.36 & 4.67 & \textbf{3.81}\\
    mass & 10.00 & 6.87 & 6.96 & \textbf{4.24}\\
    \bottomrule
\end{tabular}

\label{tbl:regression_table_model}
\vspace{-1mm}
\end{table}

\begin{table}[t]
\centering\small
\setlength{\tabcolsep}{3pt}
\caption{Physical property decoding error $D_\text{err}$ with different actions}
\vspace{-5pt}
\begin{tabular}{lccccc}
    \toprule
    & Slow Push & Sliding & Collision & 
    \tabincell{c}{Slow Push \\\& Collision} &
    \tabincell{c}{Sliding \& \\ Collision}\\
    \midrule
    friction & 8.87 & 4.07 & 6.58 & 5.92 & \textbf{3.81}\\
    mass & 9.76 & 9.23 & 7.03 & 6.75 & \textbf{4.24}\\
    \bottomrule
\end{tabular}
\label{tbl:regression_table_action}
\vspace{-5mm}
\end{table}
% \begin{figure*}[t]
% \centering
% \includegraphics[width=0.35\linewidth]{figures/friction_distance.png}
% \qquad
% \includegraphics[width=0.35\linewidth]{figures/mass_distance.png}
% \caption{\textbf{Decoding physical object properties.} The plot show the error distance on decoding object's friction coefficient and mass at each step. The plot shows that the prediction accuracy at the beginning is low, however, over the course of interactions, the average distance decreases quickly, which means our model gradually accumulates knowledge of object physics. The result also demonstrate that our model is able to generalize well to novel objects that has different shape and physical property. 
% }
% \label{fig:regression_result}
% \end{figure*}

\begin{figure}[t]
\centering
\includegraphics[width=\linewidth]{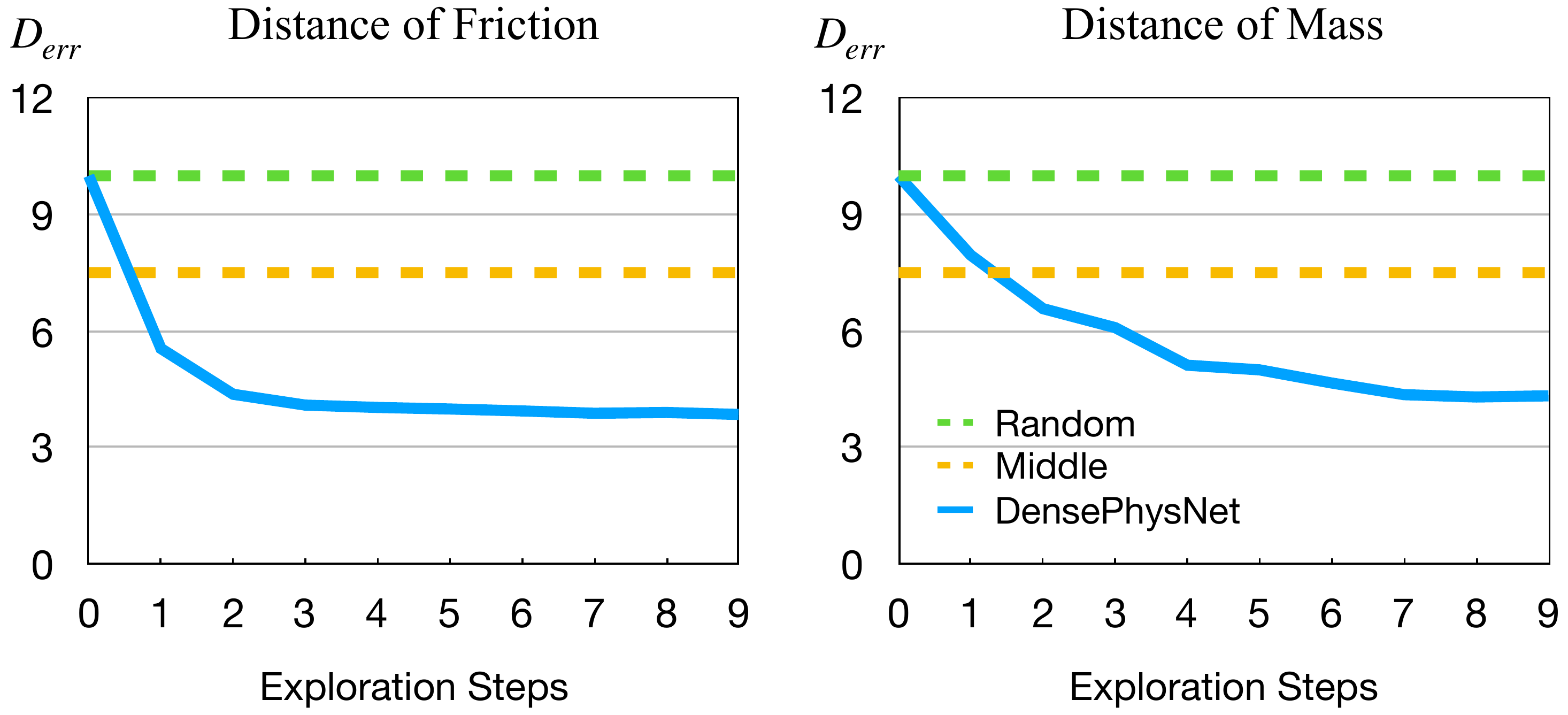}
\vspace{-20pt}
\caption{\textbf{Decoding physical object properties.} The plot shows the average distance error on decoding friction coefficient and mass at each step. At the beginning, our model's prediction accuracy is, as expected, the same as random guess; over the course of interactions, the average distance decreases quickly, which means our model gradually accumulates knowledge of object physics. 
%The result also demonstrates that our model is able to generalize well to novel objects that has different shape. 
% \shuran{can we make the text in this figure bigger? and use a pdf?}
}
\vspace{-15pt}
\label{fig:regression_result}
\end{figure}

%\mysubpara{Our model with multiple slides or collisions.} We also compare with our model trained with only sliding or collisions, but not both, to validate the importance of having multiple types of interactions.

%\mysubpara{Our model at different interaction steps.} We also show the performance of after each interaction steps. 

% \mysubpara{Single-step forward model~\cite{Agrawal2016Learning}.} We also compare our approach with the the model proposed by Agrawal~\etal~\cite{Agrawal2016Learning}. Their model consists of a forward and an inverse model, both designed to handle a single-step interaction instead of aggregating the information across multiple interactions. The forward model takes the current frame and the action as input, and predict the motion of the object for one step (see \fig{fig:application_baseline}a). 

% \mysubpara{Single-step inverse model~\cite{Agrawal2016Learning}.} Their inverse model takes the frames before and after the action as input, and predict the action parameters (see \fig{fig:application_baseline}b). 

\mypara{Results.}
\tbl{tbl:regression_table_model} shows that \model outperforms the baselines, demonstrating the importance of information aggregation across multi-step interactions in modeling physical object properties. The forward model only takes the current image and the action as input. Its representation only contains visual information, which does not help to predict physical properties. Its result is thus the same as random guessing. The inverse model takes images before and after the interaction and learns some information about physical properties from object motion. However, it cannot handle long-range data; its performance is therefore limited.

% Moreover, the comparison between different action types demonstrate that sliding helps a lot for the learning of friction, while the mass can only be inferred from the combination of sliding and collisions. Hence the diversity of action space is of vital importance. % and we believe the performance can have further improvement with more kinds of action.
\tbl{tbl:regression_table_action} demonstrates the comparison between different kinds of interactions. Compared with slow push, the performance of sliding is much better. Moreover, sliding helps a lot for the learning of friction, while the mass can only be inferred from the combination of sliding and collisions. In short, dynamic interactions are much more effective than quasi-static interactions and the diversity of action space is of vital importance.

We also conduct an ablation study to understand how the number of interaction steps affects the results. \fig{fig:regression_result} shows \model's performance in each step. For calibration, `Random' shows the performance of random guessing: $D_\text{err} = \frac{1}{30}\times\sum_{gt=1}^{30}\sum_{i=1}^{30}\frac{1}{30}\times|i-gt| \approx 10$.  `Middle' shows the performance of predicting the average mass and friction: $D_\text{err} = \frac{1}{30}\times\sum_{gt=1}^{30}|15.5-gt| = 7.5$. At the beginning, our model's prediction is similar to random guessing. Over the course of interactions, the error decreases quickly, which means our model gradually accumulates knowledge of object physics. %The result also demonstrates that our model is able to generalize well to novel objects that has different shape. 
    %\item Our model at different interaction steps: the performance of our model after each interaction step.
    
%When we push a object and it slides for a long distance, we can guess this object is very smooth. But we may also be curious about how smooth is the object? 
%Meanwhile, this model should have generalization ability to handle some new objects, which are not seen in training. %We also want to know the importance of these two actions. What if we only one kind of action?

%\mypara{Experimental Setup}
%Considering the limitation of the two baselines, 

%We train a simple linear classifier, using cross entropy as loss function. This network takes the processed representation as input and predict the probability for each category. During testing, we calculate the average distance for each piece of data:

\subsection{Application in Sliding Objects with Unknown Physics}

The ability to understand physical properties is important for flexible manipulation. In this experiment, we test whether our model can efficiently infer physical properties of unknown objects through multi-step interactions, and further, can make use of its knowledge of object physics to suggest more accurate policies in object manipulation.

\begin{figure}[!t]
\centering
\includegraphics[width=\linewidth]{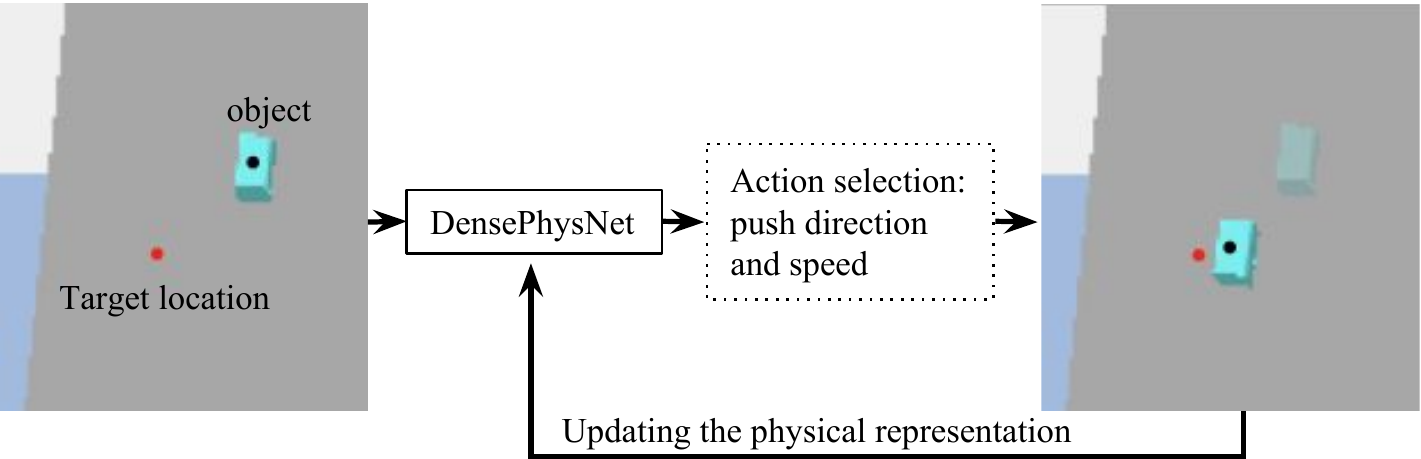}
\vspace{-15pt}
\caption{\textbf{Application in object sliding.} The goal for this task is to push an object with certain direction and speed, so that the object will slide to the target position. At each step, \model predicts the motion of the object for each possible action in the action space, then selects the one whose outcome is closest to the target. After each interaction, the new action-state representation is fed back to the multi-step aggregator to improve the object's physical representation, which consequentially improves the action predictions in the following interactions. }
\label{fig:application_push}
\vspace{-15pt}
\end{figure}

\mypara{Setup.} Our task is to push an unknown object so that it slides to the target position. The target position is uniformly randomly chosen in a circle centered on the object and with a radius of $0.25$m. 
% \jw{how, Gaussian? mean, variance? physical values, please}  
The robot needs to propose the parameters of the push, including its direction and initial speed. The sixteen possible directions are uniformly distributed on $[0, 2\pi]$ and the speed can be chosen from $\{0.96,1.28,1.44,1.6\}$rps.
% \jw{be precise.}
After each interaction, the new action-state representation is fed back to the multi-step aggregator to improve the physical representation. Intuitively, if the target object is heavy, the system should be able to infer its mass through interactions and then push it with a higher speed. \fig{fig:application_push} shows an illustration of the task.

% \jw{this is terrible. We're not writing science fictions. Everything needs to be precise and fully reproducible.}
The physical properties of the objects are randomly chosen from two distribution families:
\begin{itemize}[leftmargin=*]
    \item common objects: $m\in[0.15, 0.16]$kg and $\mu\in[0.6, 0.8]$,
    \item uncommon objects: $m\in[0.11, 0.13]$kg or $[0.18, 0.2]$kg and $\mu\in[0.4, 0.5]$ or $[0.9, 1.0]$. %\zhenjia{ugly}
\end{itemize}

We compare our model with Push-Net, and the forward and the inverse models as introduced in \sect{sec:decode}. Because our model needs a sequence of interactions to learn physical properties, we also evaluate our model after $0$, $3$, and $7$ interaction steps to understand how interaction helps control. Here, the interaction steps are just for updating the latent representations; no optical flow supervision or finetuning is needed. The policies used by different models are as follows:
\begin{itemize}[leftmargin=*]
    \item Our model and the forward model: enumerate all possible actions and predict the motion of the object, and then choose the action whose predicted position is closest to the target. 
    \item The inverse model: use the action predicted by the model from the current and the target image.
\end{itemize}
We test each model $100$ times and calculate the mean distance between the object's position after interactions and the target. 

\begin{figure}[!t]
\centering
\includegraphics[width=\linewidth]{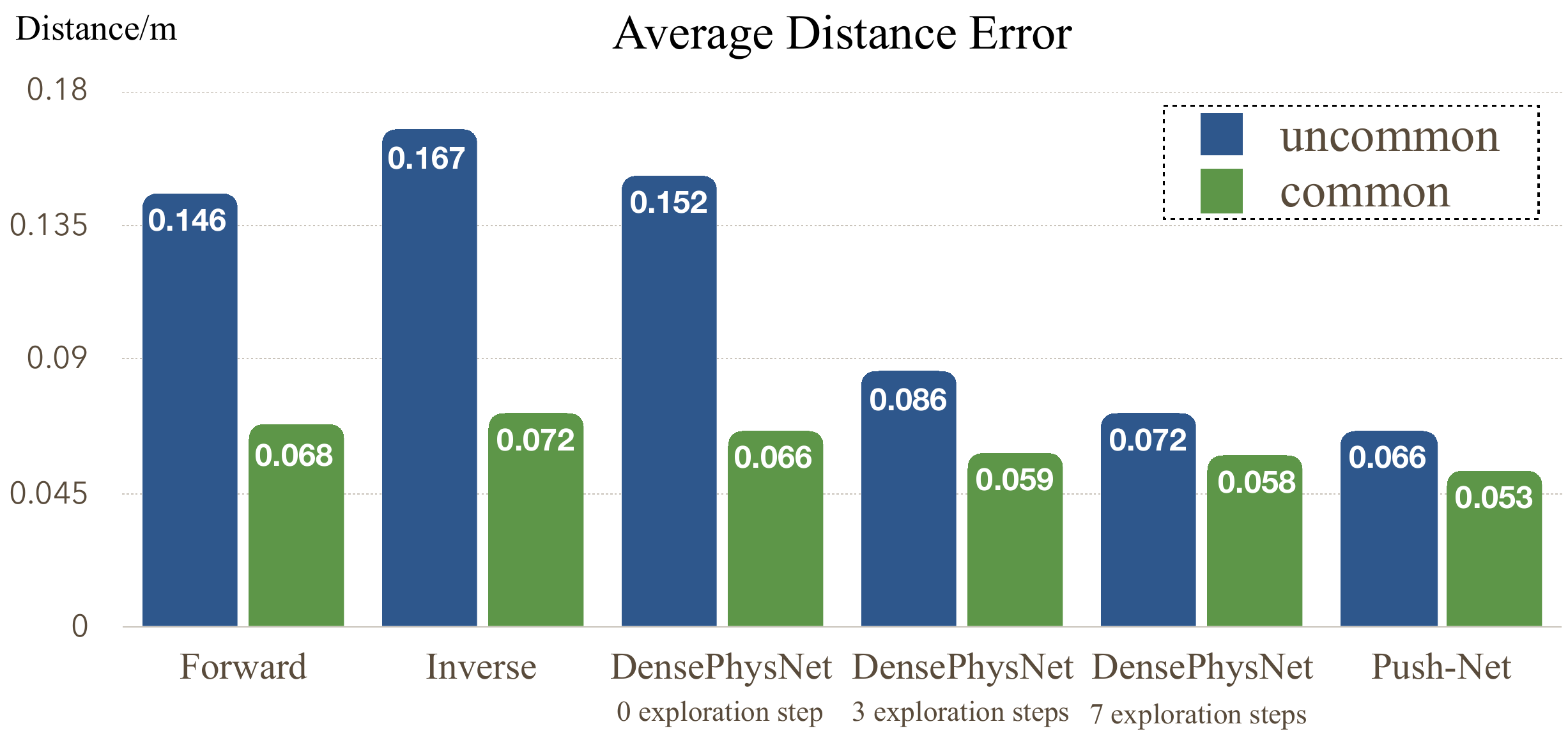}
\vspace{-20pt}
\caption{\textbf{Results on object sliding in simulation.} For objects with common physical properties, both our model and the baseline work well. However, for objects with uncommon physical properties, only our model and Push-Net (after a few exploration steps) perform well. This is because the recurrent structure in \model and Push-Net learns to infer physical properties via aggregating information from history trajectories.}
\vspace{-5pt}
\label{fig:application_pushresult}
\end{figure}

\mypara{Results.}
\fig{fig:application_pushresult} shows the results. For objects with common physical properties, all these models have similar performance with a mean distance error of $0.06$m. However, if the physical properties are uncommon, our model and Push-Net outperform the other two baselines after a few explorations. This suggests that interactions help to refine the estimation of physical properties and lead to better performance. Moreover, our model achieves similar performance with Push-Net, without requiring direct supervision as Push-Net does.
%\jw{to be honest, this experiment is weak, because baseline don't have extra exploration steps.} % The ability of the two baselines is similar to the initial state of our model. Hence, multi-step interactions can lead to accurate future prediction and further enable flexible manipulation.

\begin{figure}[!t]
\centering
\includegraphics[width=\linewidth]{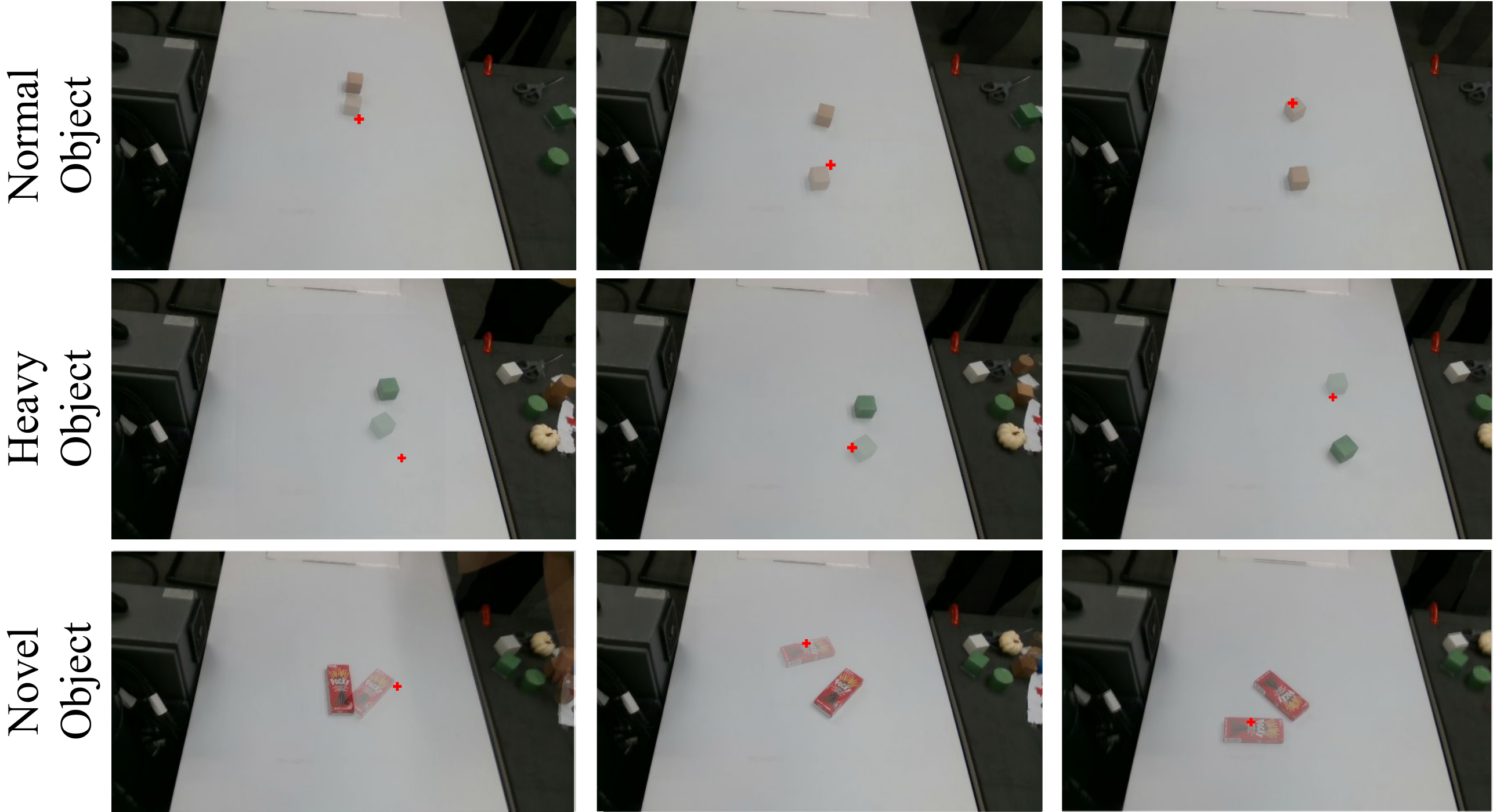}
\vspace{-10pt}
\caption{\textbf{Real-world object sliding experiment.} 
%object weight 22,72,220 kg
Each row shows the object sliding result for one object after $0$, $2$, and $5$ exploration steps, where the transparent object indicates its position after applying the suggested action. The red cross indicates the target location. The `normal' object is a wooden block that has similar physical property as the training objects in simulation. The `heavy' object is a plastic box filled with heavy metal balls. The `novel' object is a snack box that has a different shape from the training objects. }
\vspace{-5pt}
\label{fig:real_result_qualitative}
\end{figure}

\begin{figure}[!t]
\centering
\includegraphics[width=0.74\linewidth]{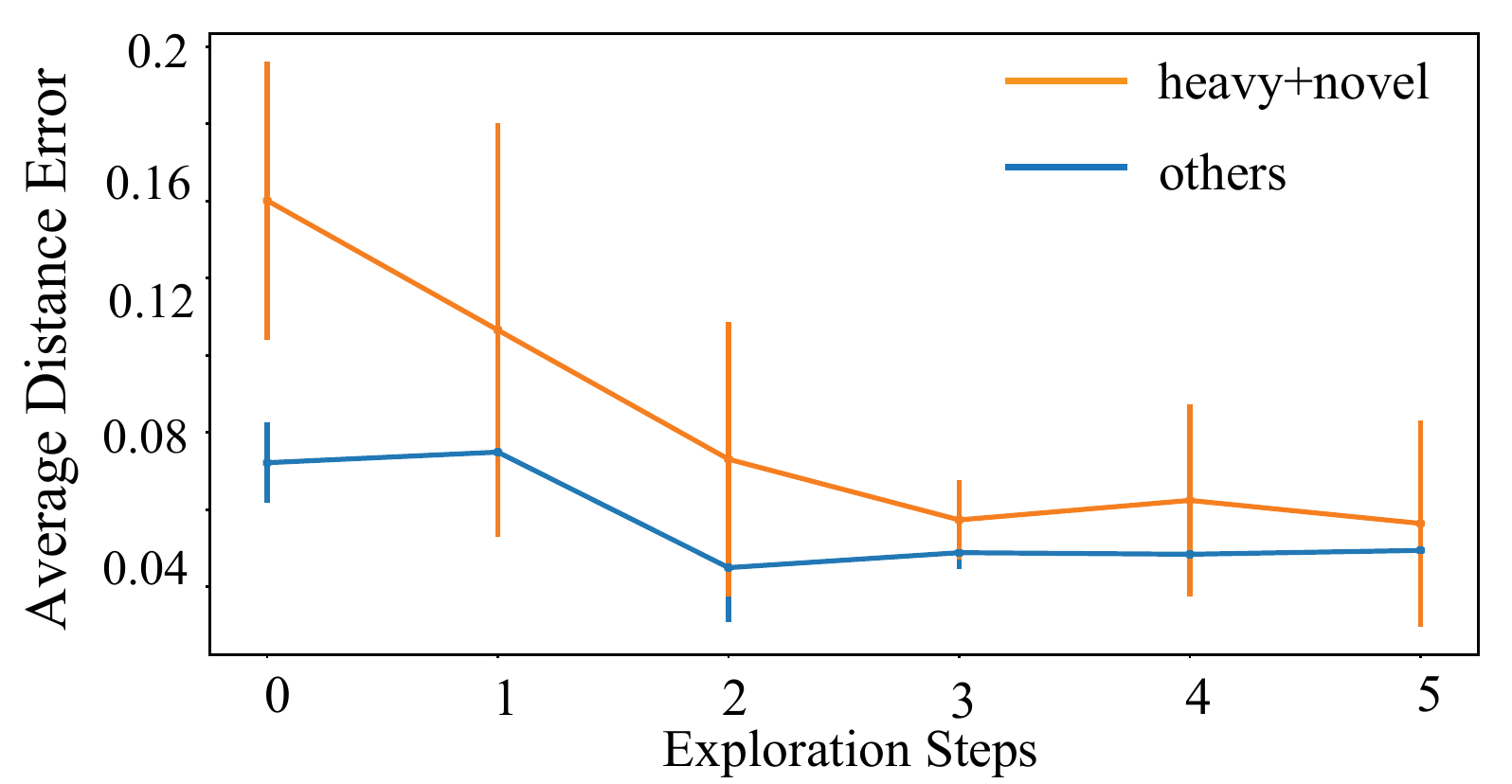}~
\includegraphics[width=0.24\linewidth]{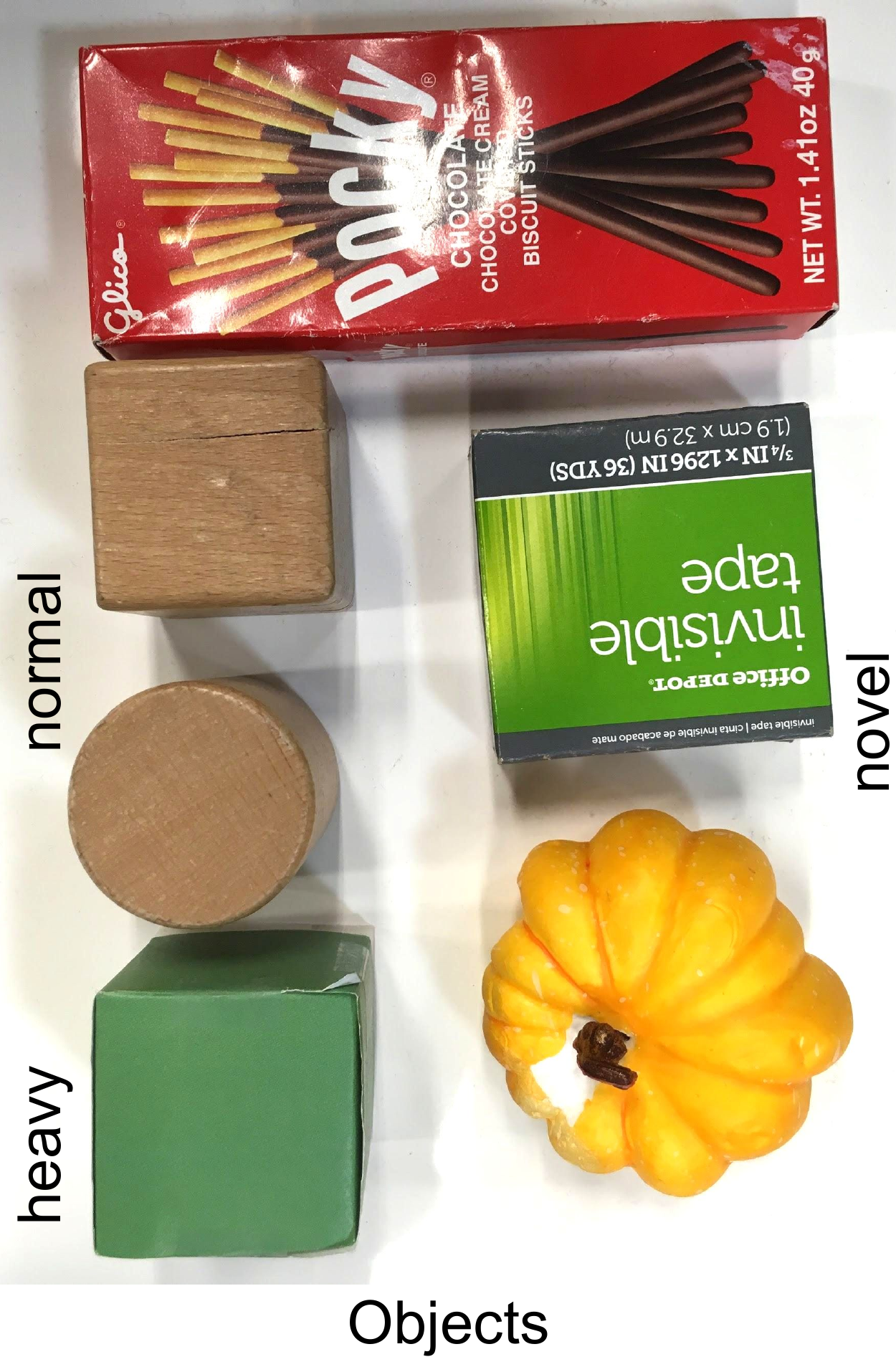}
\vspace{-15pt}
\caption{\textbf{Results on object sliding in the real world.} Our algorithm predicts more accurate actions and achieves lower errors after a few explorations steps, especially for heavy and novel objects.}
\label{fig:real_result}
\vspace{-15pt}
\end{figure}

\mypara{Real-world experiments.}
We also evaluate \model in real-world settings to push a collection of objects on a flat table. We use the model pre-trained on $8{,}000$ interaction sequences in simulation. In total, we have tested $18$ sequences for six different objects with distinct physical properties, appearance, and shape. Each sequence contains five interaction steps. The target location is manually selected for each step.  

\fig{fig:real_result_qualitative} shows qualitative results for different objects. \model not only generalizes from simulation to real-world settings, but also makes use of the exploration steps to improve its action predictions. In particular, for the objects with an unusually large mass, such as the `heavy object' in \fig{fig:real_result_qualitative}, the model without exploration (step 0) predicts a push speed that is too small for the object to slide to the target position. But just after two steps of interaction, the model learns to adjust its prediction according to the estimated physical properties. \fig{fig:real_result} shows the mean distance error for the `heavy' and `novel' objects (objects whose shape is different from those in training), as well as for all other `normal' objects. On average, the algorithm predicts more accurate actions and achieves lower errors after a few explorations steps, consistent with the experiments in simulation. As expected, the gain of exploration is more significant for heavy and novel objects than normal objects.

\subsection{Generalization}

Practical robotic systems need to generalize to complex scenarios. Here we evaluate on two cases: generalizing to scenes with more objects and generalizing to novel tasks.

\mypara{Generalizing to scenes with more objects.} 
Pixel-wise dense representations work for scenes with multiple objects, and further, generalizes to scenes with more objects than those in training scenes. In this experiment, we train the model using two objects and test it with three objects. Each scene consists of objects with different shapes and physical properties. The robot is allowed to have $19$ interactions for both training and testing. Other setups of this experiment is the same as in \sect{sec:decode}.

\begin{figure}[!t]
\centering
\includegraphics[width=\linewidth]{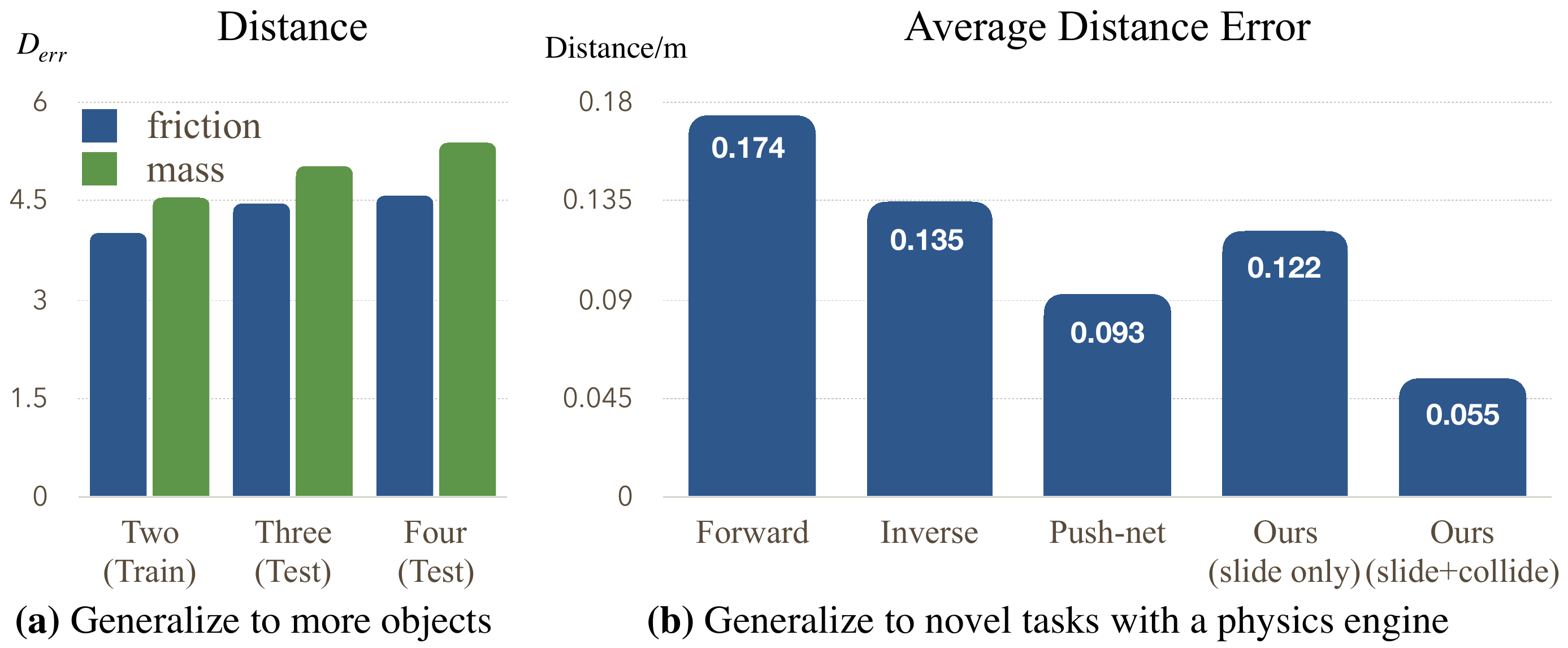}
\vspace{-20pt}
%\shuran{change or just remove "hitting the object to a target position" in the figure. It sounds just like the sliding task now. Maybe change it to "Generalize to novel tasks with physics engine"}
\caption{\textbf{Generalization.} (a) Results of generalizing to more objects. (b) Results of generalizing to a new task, where the algorithm combines the decoded object physical properties with a general physics  engine to predict planning and control parameters. In this new task, \model outperforms baselines significantly. In particular, using both types of interactions is important to achieve good generalization results.
}
\label{fig:generalization_result}
\vspace{-4mm}
\end{figure}

There are $5{,}000$ sequences with two objects for training and $1{,}500$ sequences for testing: $500$ with two objects, $500$ with three objects, and $500$ with four objects. We calculate the average distance between the predicted physical properties and ground truth for each object in each piece of data. %Considering the limitation of other baselines, here we only demonstrate the results of our \model. 
%For our model, the average distances of friction are $4.02$ (two objects), $4.48$ (three objects), and $4.59$ (four objects); as to the mass, the distances are $4.57$ (two objects), $5.04$ (three objects), and $5.41$ (four objects). 
Figure \ref{fig:generalization_result}(a) shows the average error. The small gap between scenes with different numbers objects suggests that our \model generalizes to new scenes with more objects; in contrast, baselines~\cite{Li2018Push,Agrawal2016Learning} do not have an object-wise representation and cannot directly work on scenes with multiple objects.
% \begin{table}[h]
% \centering\small
% \setlength{\tabcolsep}{6pt}
% \caption{Generalizing to scenes with more objects. \jw{can be merged with Fig 13 task generalization}}
% \vspace{-5pt}
% \begin{tabular}{l|c|cc}
%     \toprule
%     object \# & two (trained) & three & four\\
%     \midrule
%     friction &4.02 & 4.48 & 4.59 \\
%     mass & 4.57 & 5.04 & 5.41 \\
%     \bottomrule
% \end{tabular}
% \label{tbl:more_object}
% \vspace{-2mm}
% \end{table}

\mypara{Generalizing to a novel task.}
Once we have decoded the physical properties of objects, we can integrate them into a physics engine for planning and control in alternative tasks. As a demonstration, we study a new task, where the goal is to slide an auxiliary cube so that it hits the object to a target position. Here, we need to select the mass and speed of the auxiliary cube. This task is different from object collisions in the training phase: during training, the auxiliary object is a cylinder with a fixed size, mass, initial position and speed. %and slide from the same ramp with the same speed, however, here the collision objects can have different mass, friction, starting location and initial speed, it is much more flexible than the scenario in training.

The robot first interacts with the target object for $7$ steps using the pre-trained model in \sect{sec:decode}, without finetuning; %\shuran{Are these 8 step training? do you provide optical flow as supervision?} 
it then uses the decoder, also trained in \sect{sec:decode}, to decode physical properties from the physical representation. We set the target object's initial position to always be at center of the workspace, and its target position to the north of the initial position with a distance uniformly sampled from $[0.1, 0.3]$m. We set the position of the auxiliary cube to be $0.05$m south to the center. The mass of the auxiliary cube can be chosen from $\{0.4, 0.43, 0.46 \dots 0.7\}$kg and the speed can be chosen from $\{0.5, 0.53, 0.56,\dots 0.8\}$m/s, and its friction is fixed at $0.2$. 

\begin{figure}[!t]
\centering
\includegraphics[width=\linewidth]{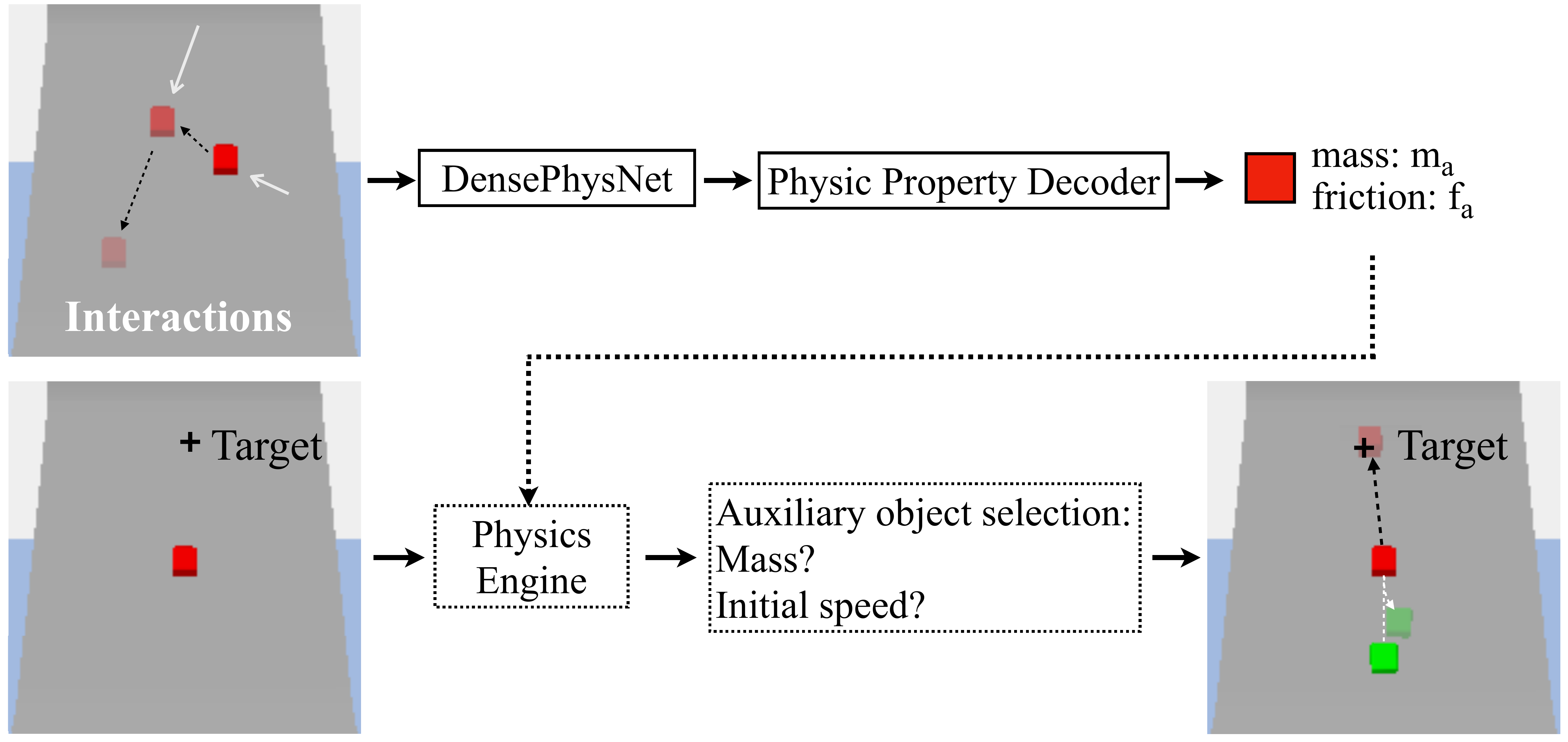}
\vspace{-20pt}
\caption{\textbf{Experiments on task generalization.} Here, the robot first interacts with the objects. The learned representation is then used to decode physical properties, which are later used in a physics engine for planning in other tasks.}
\vspace{-5pt}
\label{fig:application_collide}
\end{figure}

For each model, we simulate the collision with the decoded physical properties and each possible pair of mass and speed for the auxiliary object; we choose the pair that gives the best prediction. \fig{fig:application_collide} demonstrates the pipeline of this experiment. We test each model $100$ times and calculate the mean distance between the target position and the object’s position final position after collision. 

\fig{fig:generalization_result}(b) shows the results. Our model outperforms other baselines significantly. Further, the performance of using only sliding during exploration is not as good as the one using both sliding and collisions. Again, it demonstrates that a diverse set of action types is important to accurate prediction and better performance in control.

%\vspace{2mm}

% \subsection{Baselines:} 
% We compare our approach to other methods for representation learning (mostly visual representation learning)

% \textbf{Curious robot:} they learn to predict the action itself

% \textbf{Learning to poke by poking: }single-step exploration 

% \textbf{Deep Visual Foresight:} predicting next color frame - single action and single frame   

%We also evaluate our model with two kinds of action space used in the period of exploration. 

%At the beginning, the target object is placed in the center of workspace and the auxiliary cube is $0.05m$ south to the center. The target position is on the north side of the center and the distance is uniformly chosen from $[0.1, 0.3] (m)$.

%For each pair of mass and speed, we will set the auxiliary cube with these parameters and the target object with values predicted from the physical representation. Then the physics engine (pybullet) simulates the collision and we can get position prediction of the target object. After enumerating all pairs of parameters, we set the auxiliary cube with the one whose predicted position is closest to the target position and the target object with ground truth. Finally, we run the physics engine again and calculate the distance between the final position and the target position.
\section{Discussion and Future Work} 
\label{sec:conclusion}

We have proposed \model, a model that learns dense, physical object representations from self-supervised interactions. We have demonstrated that \model learns about object materials and physics through both qualitative and quantitative analyses. Further, we have shown that the learned representations can be used in downstream control tasks such as planar sliding to suggest more accurate action policie. Below, we discuss the key features and limitations of our design.

The design of action space is important for learning object physics. We have shown that, with both planar sliding and collisions, the model learns to infer both object mass and friction. %Infants interact with the world in many possible ways -- pushing, poking, sliding, dropping, grasping, and even tasting. 
Extending \model to accommodate a much richer set of interactions would fully demonstrate its potentials. 

\model learns physical representations from object motion via depth images; our model does not take color images as input. Color signals can however be informative of object geometry and potentially physics, as shown in prior research on visual representation learning~\cite{Pinto2016Curious,florence2018dense}; they can also help to scale our model to a richer set of objects. 
Therefore, a promising research direction is to build models that learn from both color and motion cues for better object representations.

\section*{Acknowledgments}
This work is in part supported by the Center for Brains, Minds and Machines (NSF STC award CCF-1231216), ONR MURI N00014-16-1-2007, Google, and Facebook. 

%% Use plainnat to work nicely with natbib. 

\bibliographystyle{plainnat}
\bibliography{main}

\begin{thebibliography}{34}
\providecommand{\natexlab}[1]{#1}
\providecommand{\url}[1]{\texttt{#1}}
\expandafter\ifx\csname urlstyle\endcsname\relax
  \providecommand{\doi}[1]{doi: #1}\else
  \providecommand{\doi}{doi: \begingroup \urlstyle{rm}\Url}\fi

\bibitem[Agrawal et~al.(2016)Agrawal, Nair, Abbeel, Malik, and
  Levine]{Agrawal2016Learning}
Pulkit Agrawal, Ashvin Nair, Pieter Abbeel, Jitendra Malik, and Sergey Levine.
\newblock Learning to poke by poking: Experiential learning of intuitive
  physics.
\newblock In \emph{Advances in Neural Information Processing Systems
  (NeurIPS)}, 2016.

\bibitem[Atkeson et~al.(1986)Atkeson, An, and
  Hollerbach]{atkeson1986estimation}
Christopher~G Atkeson, Chae~H An, and John~M Hollerbach.
\newblock Estimation of inertial parameters of manipulator loads and links.
\newblock \emph{The International Journal of Robotics Research (IJRR)},
  5\penalty0 (3):\penalty0 101--119, 1986.

\bibitem[Brachmann et~al.(2014)Brachmann, Krull, Michel, Gumhold, Shotton, and
  Rother]{brachmann2014learning}
Eric Brachmann, Alexander Krull, Frank Michel, Stefan Gumhold, Jamie Shotton,
  and Carsten Rother.
\newblock Learning 6d object pose estimation using 3d object coordinates.
\newblock In \emph{European Conference on Computer Vision (ECCV)}, 2014.

\bibitem[Byravan and Fox(2017)]{byravan2017se3}
Arunkumar Byravan and Dieter Fox.
\newblock Se3-nets: Learning rigid body motion using deep neural networks.
\newblock In \emph{IEEE International Conference on Robotics and Automation
  (ICRA)}. IEEE, 2017.

\bibitem[Chang et~al.(2015)Chang, Funkhouser, Guibas, Hanrahan, Huang, Li,
  Savarese, Savva, Song, Su, Xiao, Yi, and Yu]{Chang2015Shapenet}
Angel~X Chang, Thomas Funkhouser, Leonidas Guibas, Pat Hanrahan, Qixing Huang,
  Zimo Li, Silvio Savarese, Manolis Savva, Shuran Song, Hao Su, Jianxiong Xiao,
  Li~Yi, and Fisher Yu.
\newblock {Shapenet: An information-rich 3d model repository}.
\newblock \emph{arXiv:1512.03012}, 2015.

\bibitem[Choy et~al.(2016)Choy, Gwak, Savarese, and
  Chandraker]{choy2016universal}
Christopher~B Choy, JunYoung Gwak, Silvio Savarese, and Manmohan Chandraker.
\newblock Universal correspondence network.
\newblock In \emph{Advances in Neural Information Processing Systems
  (NeurIPS)}, 2016.

\bibitem[Coumans(2010)]{Coumans2010Bullet}
Erwin Coumans.
\newblock Bullet physics engine.
\newblock \emph{Open Source Software: http://bulletphysics. org}, 2010.

\bibitem[Denil et~al.(2017)Denil, Agrawal, Kulkarni, Erez, Battaglia, and
  de~Freitas]{Denil2017Learning}
Misha Denil, Pulkit Agrawal, Tejas~D Kulkarni, Tom Erez, Peter Battaglia, and
  Nando de~Freitas.
\newblock Learning to perform physics experiments via deep reinforcement
  learning.
\newblock In \emph{International Conference on Learning Representations
  (ICLR)}, 2017.

\bibitem[Diankov(2010)]{diankov_thesis}
Rosen Diankov.
\newblock \emph{Automated Construction of Robotic Manipulation Programs}.
\newblock PhD thesis, Carnegie Mellon University, Robotics Institute, 2010.

\bibitem[Ehrhardt et~al.(2017)Ehrhardt, Monszpart, Mitra, and
  Vedaldi]{Ehrhardt2017Taking}
Sebastien Ehrhardt, Aron Monszpart, Niloy Mitra, and Andrea Vedaldi.
\newblock Taking visual motion prediction to new heightfields.
\newblock \emph{arXiv:1712.09448}, 2017.

\bibitem[Finn and Levine(2017)]{Finn2017Deep}
Chelsea Finn and Sergey Levine.
\newblock Deep visual foresight for planning robot motion.
\newblock In \emph{IEEE International Conference on Robotics and Automation
  (ICRA)}, pages 2786--2793. IEEE, 2017.

\bibitem[Florence et~al.(2018)Florence, Manuelli, and
  Tedrake]{florence2018dense}
Peter~R Florence, Lucas Manuelli, and Russ Tedrake.
\newblock Dense object nets: Learning dense visual object descriptors by and
  for robotic manipulation.
\newblock In \emph{Conference on Robot Learning (CoRL)}, 2018.

\bibitem[Fragkiadaki et~al.(2016)Fragkiadaki, Agrawal, Levine, and
  Malik]{Fragkiadaki2016Learning}
Katerina Fragkiadaki, Pulkit Agrawal, Sergey Levine, and Jitendra Malik.
\newblock Learning visual predictive models of physics for playing billiards.
\newblock In \emph{International Conference on Learning Representations
  (ICLR)}, 2016.

\bibitem[He et~al.(2016)He, Zhang, Ren, and Sun]{He2016Deep}
Kaiming He, Xiangyu Zhang, Shaoqing Ren, and Jian Sun.
\newblock Deep residual learning for image recognition.
\newblock In \emph{IEEE Conference on Computer Vision and Pattern Recognition
  (CVPR)}, 2016.

\bibitem[Kingma and Ba(2015)]{Kingma2015Adam}
Diederik~P. Kingma and Jimmy Ba.
\newblock Adam: A method for stochastic optimization.
\newblock In \emph{International Conference on Learning Representations
  (ICLR)}, 2015.

\bibitem[Li et~al.(2018)Li, Hsu, and Lee]{Li2018Push}
Jue~Kun Li, David Hsu, and Wee~Sun Lee.
\newblock Push-net : Deep planar pushing for objects with unknown physical
  properties.
\newblock In \emph{Robotics: Science and Systems (RSS)}, 2018.

\bibitem[Maaten and Hinton(2008)]{maaten2008visualizing}
Laurens van~der Maaten and Geoffrey Hinton.
\newblock Visualizing data using t-sne.
\newblock \emph{Journal of Machine Learning Research (JMLR)}, 9:\penalty0
  2579--2605, 2008.

\bibitem[Paszke et~al.(2017)Paszke, Gross, Chintala, Chanan, Yang, DeVito, Lin,
  Desmaison, Antiga, and Lerer]{Paszke2017Automatic}
Adam Paszke, Sam Gross, Soumith Chintala, Gregory Chanan, Edward Yang, Zachary
  DeVito, Zeming Lin, Alban Desmaison, Luca Antiga, and Adam Lerer.
\newblock Automatic differentiation in pytorch.
\newblock In \emph{Advances in Neural Information Processing Systems (NeurIPS)
  Workshops}, 2017.

\bibitem[Pinto and Gupta(2017)]{pinto2017learning}
Lerrel Pinto and Abhinav Gupta.
\newblock Learning to push by grasping: Using multiple tasks for effective
  learning.
\newblock In \emph{IEEE International Conference on Robotics and Automation
  (ICRA)}, 2017.

\bibitem[Pinto et~al.(2016)Pinto, Gandhi, Han, Park, and
  Gupta]{Pinto2016Curious}
Lerrel Pinto, Dhiraj Gandhi, Yuanfeng Han, Yong-Lae Park, and Abhinav Gupta.
\newblock The curious robot: Learning visual representations via physical
  interactions.
\newblock In \emph{European Conference on Computer Vision (ECCV)}, 2016.

\bibitem[Schmidt et~al.(2017)Schmidt, Newcombe, and Fox]{schmidt2017self}
Tanner Schmidt, Richard Newcombe, and Dieter Fox.
\newblock Self-supervised visual descriptor learning for dense correspondence.
\newblock \emph{IEEE Robotics and Automation Letters (RA-L)}, 2\penalty0
  (2):\penalty0 420--427, 2017.

\bibitem[Shotton et~al.(2013)Shotton, Glocker, Zach, Izadi, Criminisi, and
  Fitzgibbon]{shotton2013scene}
Jamie Shotton, Ben Glocker, Christopher Zach, Shahram Izadi, Antonio Criminisi,
  and Andrew Fitzgibbon.
\newblock Scene coordinate regression forests for camera relocalization in
  rgb-d images.
\newblock In \emph{IEEE Conference on Computer Vision and Pattern Recognition
  (CVPR)}, 2013.

\bibitem[Thewlis et~al.(2017)Thewlis, Bilen, and
  Vedaldi]{thewlis2017unsupervised}
James Thewlis, Hakan Bilen, and Andrea Vedaldi.
\newblock Unsupervised learning of object frames by dense equivariant image
  labelling.
\newblock In \emph{Advances in Neural Information Processing Systems
  (NeurIPS)}, 2017.

\bibitem[Watters et~al.(2017)Watters, Tacchetti, Weber, Pascanu, Battaglia, and
  Zoran]{Watters2017Visual}
Nicholas Watters, Andrea Tacchetti, Theophane Weber, Razvan Pascanu, Peter
  Battaglia, and Daniel Zoran.
\newblock Visual interaction networks.
\newblock In \emph{Advances in Neural Information Processing Systems
  (NeurIPS)}, 2017.

\bibitem[Wu et~al.(2015)Wu, Yildirim, Lim, Freeman, and
  Tenenbaum]{Wu2015Galileo}
Jiajun Wu, Ilker Yildirim, Joseph~J Lim, William~T Freeman, and Joshua~B
  Tenenbaum.
\newblock Galileo: Perceiving physical object properties by integrating a
  physics engine with deep learning.
\newblock In \emph{Advances in Neural Information Processing Systems
  (NeurIPS)}, 2015.

\bibitem[Wu et~al.(2017)Wu, Lu, Kohli, Freeman, and Tenenbaum]{Wu2017Learning}
Jiajun Wu, Erika Lu, Pushmeet Kohli, William~T Freeman, and Joshua~B Tenenbaum.
\newblock Learning to see physics via visual de-animation.
\newblock In \emph{Advances in Neural Information Processing Systems
  (NeurIPS)}, 2017.

\bibitem[Xue et~al.(2016)Xue, Wu, Bouman, and Freeman]{Xue2016Visual}
Tianfan Xue, Jiajun Wu, Katherine Bouman, and William~T Freeman.
\newblock Visual dynamics: Probabilistic future frame synthesis via cross
  convolutional networks.
\newblock In \emph{Advances in Neural Information Processing Systems
  (NeurIPS)}, 2016.

\bibitem[Ye et~al.(2018)Ye, Wang, Davidson, and Gupta]{ye2018interpretable}
Tian Ye, Xiaolong Wang, James Davidson, and Abhinav Gupta.
\newblock Interpretable intuitive physics model.
\newblock In \emph{European Conference on Computer Vision (ECCV)}, 2018.

\bibitem[Yu et~al.(2017)Yu, Tan, Liu, and Turk]{yu2017preparing}
Wenhao Yu, Jie Tan, C~Karen Liu, and Greg Turk.
\newblock Preparing for the unknown: Learning a universal policy with online
  system identification.
\newblock \emph{arXiv preprint arXiv:1702.02453}, 2017.

\bibitem[Yu et~al.(2005)Yu, Arima, and Tsujio]{yu2005estimation}
Yong Yu, Tetsu Arima, and Showzow Tsujio.
\newblock Estimation of object inertia parameters on robot pushing operation.
\newblock In \emph{IEEE International Conference on Robotics and Automation
  (ICRA)}, 2005.

\bibitem[Zeng et~al.(2017)Zeng, Song, Nie{\ss}ner, Fisher, and
  Xiao]{Zeng20173DMatch}
Andy Zeng, Shuran Song, Matthias Nie{\ss}ner, Matthew Fisher, and Jianxiong
  Xiao.
\newblock 3dmatch: Learning the matching of local 3d geometry in range scans.
\newblock In \emph{IEEE Conference on Computer Vision and Pattern Recognition
  (CVPR)}, 2017.

\bibitem[Zeng et~al.(2018)Zeng, Song, Welker, Lee, Rodriguez, and
  Funkhouser]{zeng2018learning}
Andy Zeng, Shuran Song, Stefan Welker, Johnny Lee, Alberto Rodriguez, and
  Thomas Funkhouser.
\newblock Learning synergies between pushing and grasping with self-supervised
  deep reinforcement learning.
\newblock In \emph{IEEE International Conference on Intelligent Robots and
  Systems (IROS)}, 2018.

\bibitem[Zheng et~al.(2018)Zheng, Luo, Wu, and
  Tenenbaum]{Zheng2018Unsupervised}
David Zheng, Vinson Luo, Jiajun Wu, and Joshua~B Tenenbaum.
\newblock Unsupervised learning of latent physical properties using
  perception-prediction networks.
\newblock In \emph{Conference on Uncertainty in Artificial Intelligence (UAI)},
  2018.

\bibitem[Zhou et~al.(2019)Zhou, Pinto, and Gupta]{zhou2018environment}
Wenxuan Zhou, Lerrel Pinto, and Abhinav Gupta.
\newblock Environment probing interaction policies.
\newblock In \emph{International Conference on Learning Representations
  (ICLR)}, 2019.

\end{thebibliography}

\end{document}